\documentclass{article}

\PassOptionsToPackage{numbers,compress}{natbib}

\usepackage{graphicx}

    \usepackage[preprint]{neurips_2025}

\usepackage[utf8]{inputenc} 
\usepackage[T1]{fontenc}    
\usepackage{hyperref}       
\usepackage{url}            
\usepackage{booktabs}       
\usepackage{amsfonts}       
\usepackage{nicefrac}       
\usepackage{microtype}      
\usepackage{subcaption}
\usepackage{xcolor}         
\usepackage{amsmath}
\usepackage{enumitem}
\usepackage{wrapfig}
\usepackage{adjustbox}
\usepackage{xcolor}
\usepackage{colortbl}
\usepackage{graphicx}
\usepackage{multicol}
\usepackage{multirow}
\usepackage[capitalize,noabbrev]{cleveref}

\title{Multimodal Conditional Information Bottleneck for Generalizable AI-Generated Image Detection}

\author{%
  Haotian Qin$^1$, Dongliang Chang$^{1\dag}$, Yueying Gao$^1$, Bingyao Yu$^2$, Lei Chen$^2$ and Zhanyu Ma$^1$\\ 
$^1$\textit{Beijing University of Posts and Telecommunications, Beijing, China}\\
$^2$\textit{Tsinghua University, Beijing, China}\\
}
\begin{document}
\maketitle
\let\thefootnote\relax\footnotetext{$\dag$ Corresponding author}
\begin{abstract}
    Although existing CLIP-based methods for detecting AI-generated images have achieved promising results, they are still limited by severe feature redundancy, which hinders their generalization ability. To address this issue, incorporating an information bottleneck network into the task presents a straightforward solution. However, relying solely on image-corresponding prompts results in suboptimal performance due to the inherent diversity of prompts.
    In this paper, we propose a multimodal conditional bottleneck network to reduce feature redundancy while enhancing the discriminative power of features extracted by CLIP, thereby improving the model's generalization ability. We begin with a semantic analysis experiment, where we observe that arbitrary text features exhibit lower cosine similarity with real image features than with fake image features in the CLIP feature space—a phenomenon we refer to as ``bias''.
    Therefore, we introduce InfoFD, a text-guided AI-generated image detection framework. InfoFD consists of two key components: the Text-Guided Conditional Information Bottleneck (TGCIB) and Dynamic Text Orthogonalization (DTO). TGCIB improves the generalizability of learned representations by conditioning on both text and class modalities. DTO dynamically updates weighted text features, preserving semantic information while leveraging the global ``bias''. Our model achieves exceptional generalization performance on the GenImage dataset and latest generative models.
    Our code is available at \href{https://github.com/Ant0ny44/InfoFD}{InfoFD}.
\end{abstract}
\section{Introduction}
\label{sec:1}
\begin{figure}[ht]
    \begin{subfigure}[t]{0.58\textwidth}
        \centering
        \includegraphics[width=\columnwidth]{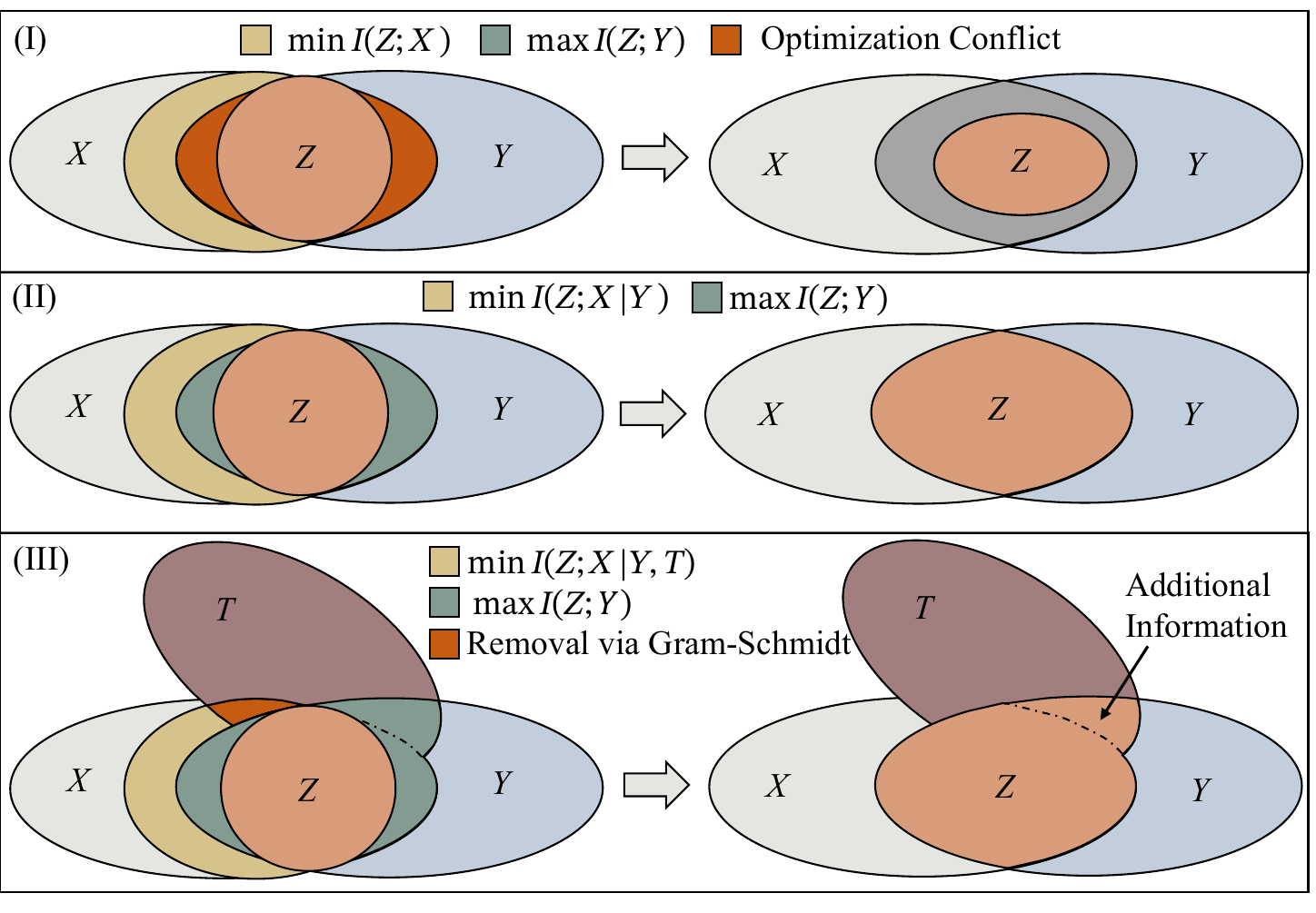}
        \caption{\textbf{Comparison of Different Information Bottleneck Methods. }}
        \label{fig:com_mmib}
    \end{subfigure}
    \hfill
    \begin{subfigure}[t]{0.38\textwidth}
        \centering
        \rotatebox{90}{\includegraphics[width=\columnwidth]{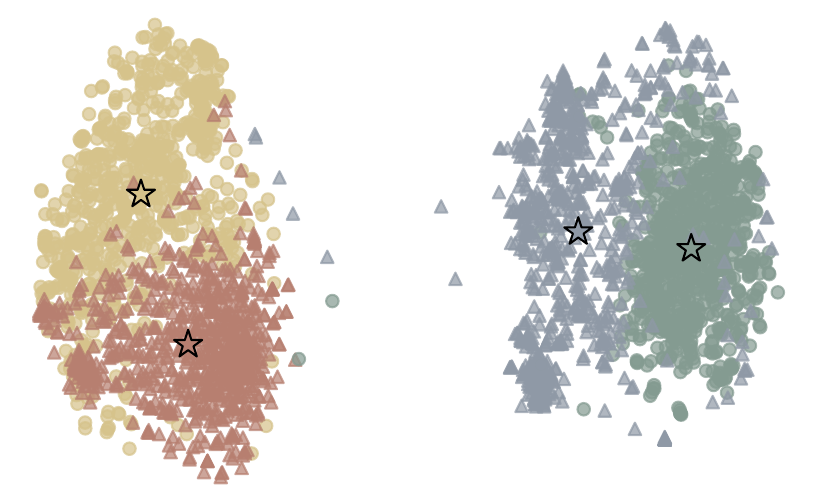}}
        \caption{\textbf{T-SNE visualization of ``bias''.}}
        \label{fig:intro_bias_vis}
    \end{subfigure}
    \caption{\textbf{Illustration of Multimodal Conditional Information Bottleneck and T-SNE visualization of ``bias''.} \textbf{(a)} \textbf{(I)} The standard Information Bottleneck (IB) \cite{vib2017}. \textbf{ (II)} The Conditional Information Bottleneck (CIB) \cite{cib2020}. \textbf{(III)} Our MultiModal Conditional Information Bottleneck (MMCIB). \textbf{ (b) } \textcolor{blue}{Blue triangles}: fake images. \textcolor{green}{Green circles}: real images. \textcolor{red}{Red triangles}: prompts for fake images. \textcolor{yellow}{Yellow circles}: prompts for real images. Image features are obtained from the [CLS] output of CLIP's layer11 followed by projection through the final layer, while text features are encoded using CLIP's text encoder. The real images and their corresponding text are sourced from ImageNet \cite{imagenet2009}, while the fake images and their corresponding text are obtained from Stable Diffusion \cite{diffusiondb2022}. The centroids corresponding to each category are marked with pentagrams. }
    \vskip -0.2in
\end{figure}
The continuous advancement of image generation models has led to increasingly realistic fake images, raising significant societal concerns \cite{fatformer2024}, such as the erosion of trust in media due to the proliferation of fake news and misinformation. 
Moreover, fake images generated by different types of generative models often exhibit distinct characteristics \cite{cliping2024}, which leads to suboptimal performance of some detection methods when confronted with unseen fake images \cite{unifd2023}. 
Consequently, there is an urgent need for a generalized approach to detecting generated images.

A prevalent approach in current research involves leveraging the powerful feature extraction capabilities of pre-trained models to obtain generalizable features for AI-generated images \cite{unifd2023,lasted2023,raise2024,fatformer2024,aide2024}, with CLIP \cite{clip2021,cliping2024} being particularly widely adopted due to its exceptional generalization capability.
However, while CLIP's features are highly expressive, they may exhibit redundancy \cite{notallfeatures2023,closer2025}.
Fortunately, the information bottleneck method \cite{vib2017} provides a robust theoretical and practical framework to address this redundancy, enabling the extraction of compressed and generalizable representations. Given the input $X$, the target prediction $Y$, traditional information bottleneck minimize $I(Z; X)$ to compress representations $Z$ for generalizability.

However, although traditional information bottleneck algorithms achieve generalization through compression, they may inadvertently discard critical information during optimization that should ideally be preserved \cite{cib2020}.
\citet{cib2020} reported this problem in their work, attributing it to the optimization conflict between $min.  I(Z;X)$  and  $max. I(Z;Y)$ , as illustrated in \cref{fig:com_mmib}(I). The CIB can resolve this conflict, thereby preventing the loss of beneficial information during optimization (\cref{fig:com_mmib}(II)). Moreover, inspired by recent work \cite{lasted2023,defake2023,fatformer2024}, we note that incorporating textual modality can enhance CLIP's ability to detect AI-generated images. Building on this foundation, we aim to design an information bottleneck method capable of leveraging multimodal information for compression.  
We seek to incorporate the text modality $T$ as a condition, allowing the model to learn representations $Z$ that include additional information about $Y$, as depicted in \cref{fig:com_mmib} (III). 
However, experimental results indicate that using only corresponding prompts as conditions leads to suboptimal performance. We hypothesize that this is due to the substantial variability in prompts across different images, which hinders model convergence.


To resolve this issue, we conducted textual analysis experiments.
We first discard the semantic features of the text by encoding random text sequences using CLIP's text encoder to derive random text features and calculate cosine similarity between the average of the random text features and those of real and fake images from various models. 
From these experiments, we conclude:  
(i) In the CLIP feature space, the cosine similarity between text features and real images is lower than that with fake images, regardless of whether the text is image descriptions or random text, as illustrated in \ref{fig:intro_bias_vis} and \cref{fig:diff_datasets_random_texts}.
We term this phenomenon ``bias''.  
(ii) For CLIP's visual encoder, the difference in cosine similarity between visual features and text features varies across layers for real and fake images, with the largest difference observed at layer 11.

Therefore, we propose InfoFD, a multimodal conditional information bottleneck AI-generated image detector. 
InfoFD comprises two core components: the \textbf{T}ext-\textbf{G}uided \textbf{C}onditional \textbf{I}nformation \textbf{B}ottleneck (\textbf{TGCIB}) module and the \textbf{D}ynamic \textbf{T}ext \textbf{O}rthogonalization (\textbf{DTO}) module. 
Unlike prior works \cite{mib2022,dmib2024,EnhancingMultimodalEntity2024}, TGCIB conditions on both additional modality information $ T $ and class labels $ Y $, integrating them via Gram-Schmidt orthogonalization, enabling the model to learn additional representational information during training without requiring $ T $ or $ Y $ during inference, as illustrated in \cref{fig:com_mmib} (III).
Additionally, to enhance generalization, we introduce \textbf{C}omposite \textbf{G}aussian \textbf{P}erturbation (\textbf{CGP}) to augment CLIP's image features, simulating domain variations across generative models. 
Inspired by the memory bank, the DTO module dynamically weights and averages text features within each batch using historical text features during training, balancing local mini-batch semantics and global bias to guide image feature extraction. 
The orthogonalized balanced text features from DTO then serve as conditional inputs to optimize the compressed representations of image features. 
Finally, we evaluated our method on GenImage \cite{genimage2024} and latest generative models \cite{cospy2025}, showcasing its generalization across diverse generated images.
Our contributions can be summarized as follows:
 
\begin{enumerate}[nosep]
    \item We identified CLIP's ``bias'', in the CLIP feature space, the cosine similarity between any text features and fake images is higher than that with real images.
     
    \item We proposed a multimodal conditional information bottleneck method, which leverages information from another modality during training to enhance model performance.

    \item We proposed a SOTA AI-generated image detector that outperforms existing approaches without CLIP fine-tuning.
\end{enumerate}  

\section{Related Work}
\label{sec:2}

\subsection{AI-generated Image Detection Models}
Generative Adversarial Networks (GANs) \cite{biggan2018, stylegan2019, stylegan22020,stylegan32021, styleganxl2022} and Diffusion models \cite{ddpm2020,ddim2021,ldm2022,controlnet2023,sdxl2024} are currently the predominant methods for generating forged images. 
Due to the distinct characteristics of images produced by different generative models \cite{unifd2023,fatformer2024}, most existing AI-generated image detection models struggle to accurately discern unseen images.
Current AI-generated image detection models can be categorized into image-based methods \cite{cnn2020,FF++2019,facegao2023,properties2023}, frequency domain-based methods \cite{lfa2020,frequency2021}, reconstruction-based methods \cite{dire2023,drct2024,watermark2024}, and pretrained model-based methods \cite{unifd2023,fatformer2024,raise2024,llmgao2025}.

In pretrained-based methods, CLIP, a large-scale multimodal pre-trained model based on contrastive learning, proposed by Radford et al. \cite{clip2021}, exhibits remarkable generalizability.
CLIP-based AI-generated image detection models fall into two categories: single-modal detection \cite{unifd2023} and multimodal detection \cite{defake2023,lasted2023,raise2024,fatformer2024}. 
Single-modal methods, such as UniFD \cite{unifd2023}, use CLIP's image encoder to extract generalized image features for detection. 
While simple and effective, these methods underutilize CLIP's powerful multimodal feature space.
Multimodal methods leverage the CLIP text encoder to incorporate information from an additional modality, achieving superior classification performance. 
For instance, Fatformer \cite{fatformer2024} uses an LGA module to align local image patches with global text embeddings, directing the image encoder to focus on forgery-related features. 
LASTED \cite{lasted2023} designs dataset-specific prompts paired with images to fine-tune CLIP, while C2P-CLIP \cite{c2p2025} clusters the forgery-related features extracted by CLIP and injects the cluster centers into CLIP as prior knowledge, as prompts.
Although multi-modal models effectively utilize CLIP's text features, LASTED achieves comparable accuracy even with non-semantic text inputs, motivating our investigation into why text features enhance AI-generated image detection.

\subsection{Multimodal Information Bottleneck}

The Variational Information Bottleneck (VIB) employs variational approximations to bound the optimization objective of the information bottleneck, allowing the IB objective to be integrated into the optimization framework of deep learning \cite{vib2017}, thereby applying the information bottleneck to the field of deep learning.
Owing to its applicability and efficacy in enhancing the generalization of compressed representations \cite{emergence2018}, the VIB has been extensively utilized in computer vision \cite{ibcv2019} and natural language processing \cite{ibnlp2020} tasks.
In the realm of multimodal research, the VIB is commonly employed to address issues related to information fusion across different modalities \cite{ibmsa2022,unpairedmmib2019}. 
Wang et al. \cite{clipib2023} enhanced the interpretability of CLIP through the information bottleneck method.
Fang et al. \cite{dmib2024} designed a dynamic fusion method based on the information bottleneck to filter redundant data. 
Cui et al. \cite{EnhancingMultimodalEntity2024} mitigated modality noise and gaps by applying IB separately to both modalities before fusion.
The Conditional Information Bottleneck (CIB) \cite{cib2020} introduces conditional information of $Y$ on top of VIB, thereby refining the constraints of the IB.
Choi and Lee \cite{tsicib2023} proposed in their work the use of CIB to reduce redundant temporal context information, addressing the substantial loss of temporal dependencies that occurs when applying IB directly to imputation problems.
Lee et al. \cite{graphCIB2023} improved the learning of intermolecular relationships by introducing a conditional graph information bottleneck, which allows the core substructures to adapt their interactions with other molecules, thereby simulating the essence of chemical reactions.
These works lack consideration for multi-modal conditional information, which is precisely one of the issues we aim to address. 
By ingeniously incorporating additional modality information ($T$) alongside target information ($Y$) as conditions, we propose a multi-modal conditional information bottleneck method.
\section{Preliminaries}
\label{sec:preliminaries}

From an intuitive perspective, textual features provide semantic information that helps CLIP's image encoder focus on regions with forgery characteristics. 
Additionally, most works \cite{fatformer2024,lasted2023,cliping2024} utilize the text modality through fine-tuning CLIP, raising two key questions:
However, Wu et al. \cite{lasted2023} found that discarding semantic information and using random characters as prompts yielded nearly identical results, challenging the necessity of semantic information. 
\label{sec:two_questions}
\begin{enumerate}[nosep]
    \item \textbf{Why do text features remain effective even without semantic information?}

    \item \textbf{Can we directly leverage CLIP's original feature space to utilize the text modality?}
\end{enumerate}

\begin{figure}[ht]
    \begin{subfigure}[ht]{0.47\linewidth} 
        \centering
        \includegraphics[width=\linewidth]{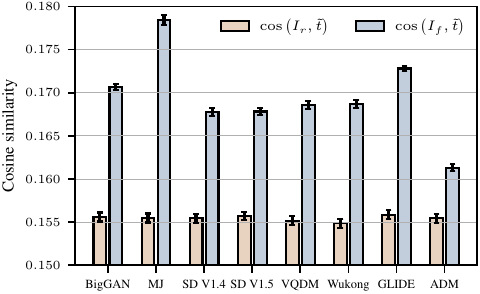}
        \vskip -0.05in
        \caption{}
        \label{fig:diff_datasets_random_texts}
    \end{subfigure}
    \begin{subfigure}[ht]{0.49\linewidth}
        \centering
        \includegraphics[width=\linewidth]{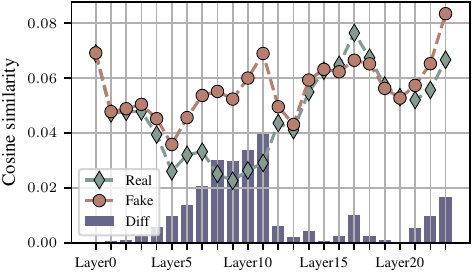}
        \vskip -0.05in
        \caption{}
        \label{fig:layer_exp_in_pre}
    \end{subfigure}
    \vskip -0.1in
    \caption{\textbf{(a)} \textbf{Mean $cos (\cdot, \cdot)$ of different datasets with the same pooling random text features $ \tilde{t} $}. Here, $I_r$ denotes real image features encoded by layer11,  $I_f$ denotes fake image features, and $ \tilde{t} $ consists of 1,000 randomly generated strings with lengths between 30 and 70, composed of randomly mixed uppercase and lowercase letters, using $3-\sigma$ error bars. All images are sourced from ImageNet \cite{imagenet2009}. For more detailed experimental results, please refer to \cref{sec:appendix_exp_preliminaries}.
            \textbf{(b)} \textbf{Outputs of different blocks in CLIP's ViT encoder.} Real images from ImageNet and fake images from Stable Diffusion V1.5 in GenImage, with 2000 images each.}
\end{figure}

\label{sec:analysis}
To address these questions, we explore the role of text features without semantic information in AI-generated image detection. 
For simplicity, $\cos(\cdot, \cdot)$ denotes cosine similarity, computed using randomly generated character sequences. 
We extend this to AI-generated images from various models, with results in \cref{fig:diff_datasets_random_texts}.

Surprisingly, the results in \cref{fig:diff_datasets_random_texts} demonstrate that even when the text features lack any semantic information, the text features still can contribute to forgery detection.
We further analyzed the impact of different CLIP layers' [CLS] outputs on this phenomenon and found it most pronounced in layer11's output. Consequently, we considered utilizing layer11 as feature extractor, as illustrated in \cref{fig:layer_exp_in_pre}. 

This phenomenon is observed across images generated by different models, which inspires us to develop a generalizable AI-generated image detection model based on this property. The finding addresses the two questions posed in \cref{sec:two_questions}. Furthermore, we conducted this experiment on more generative models \cite{genimage2024, aide2024} and obtained consistent conclusions. Detailed experimental results can be found in the \cref{sec:appendix_exp_preliminaries}.

Observing this phenomenon, we conclude that \textbf{CLIP text features exhibit a ``bias'' toward image features, demonstrating discriminative ability between fake and real images in the CLIP text feature space.} Even with meaningless input text, the encoded vectors aid in classifying real and fake images, while semantic features of the encoded prompts further enhance this effect. Moreover, based on the results from different datasets in \cref{fig:diff_datasets_random_texts}, CLIP's ``bias'' exhibits generalization capabilities. 

\section{Method}
\label{sec:4}

\begin{figure}[ht]
    \centerline{\includegraphics[width=1.0\columnwidth]{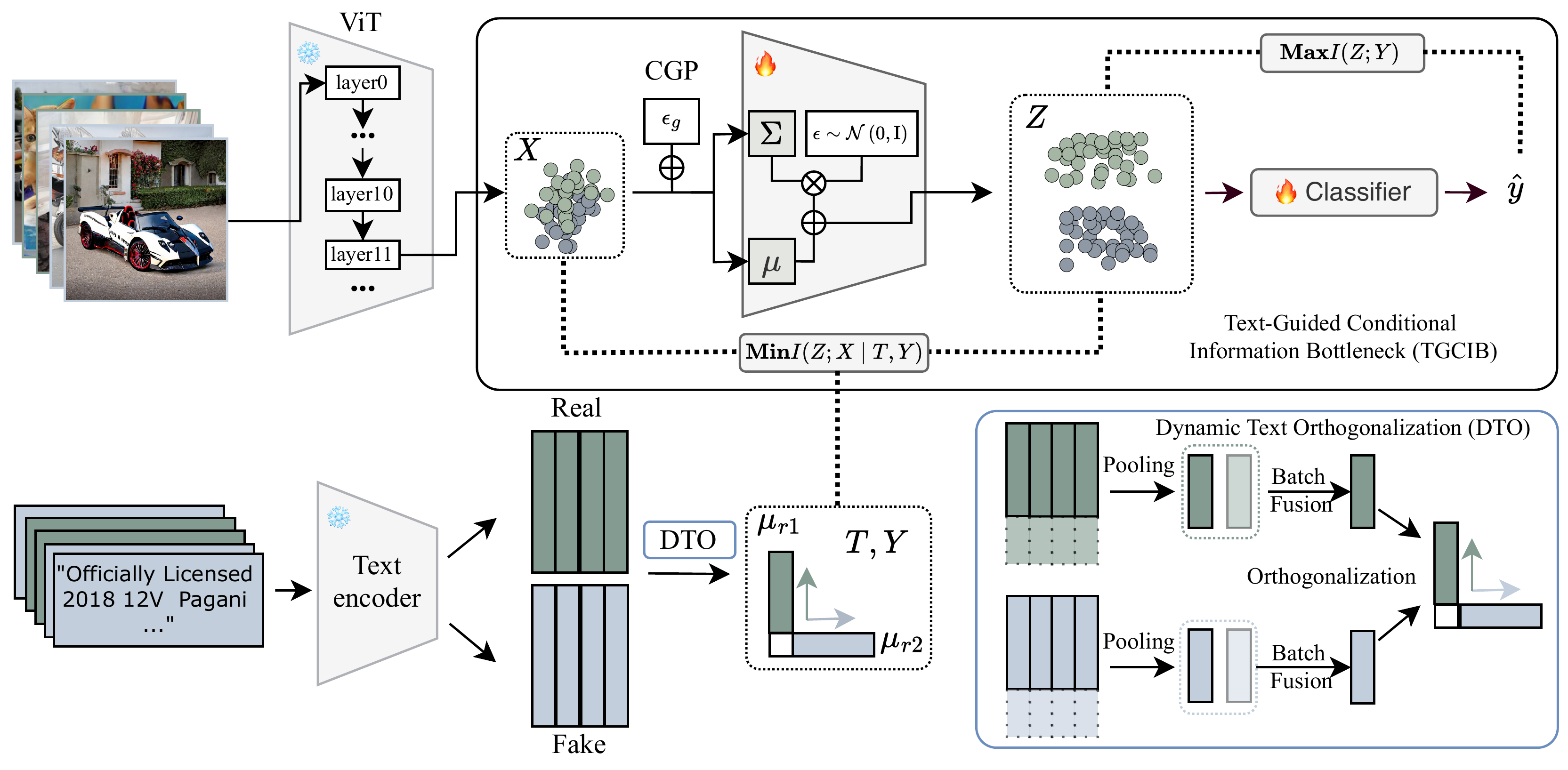}}
    \caption{\textbf{Pipeline of infoFD.} InfoFD is composed of two components: prompts processing and image processing. 
    During the text processing phase, the texts are first categorized into real and fake groups based on their corresponding image classes. 
    \textbf{DTO:} Subsequently, we employ DTO for information integration, ultimately obtaining $ r(z \mid t, y) $.
    We directly utilize the first 12 layers of the ViT in CLIP as the image encoder, based on our findings in the Preliminaries section. 
    \textbf{TGCIB:} After being encoded by CLIP's ViT, the image features are augmented using Composite Gaussian Perturbation (CGP) and then used reparameterization to simulate sampling and $ r(z \mid t, y) $ to guide optimization. Here, $\epsilon_g$  is composite noise generated by combining Gaussian and uniform distributions.}
    \label{fig:pipeline}
    \vskip -0.1in
\end{figure}
In this section, we present a multimodal conditional information bottleneck framework which allows the model to leverage information from an additional modality for training, then we detail our AI-generated image detection model, namely InfoFD. 
To the best of our knowledge, infoFD is the first approach to leverage multimodal information for auxiliary training, based on the conditional information bottleneck. Our pipeline is shown in \cref{fig:pipeline}.

\subsection{Multimodal Conditional Information Bottleneck}
\label{mmcib}
\paragraph{Learning objective.} As in previous work \cite{vib2017,cib2020,IS2021,EnhancingMultimodalEntity2024}, our goal is to obtain a latent representation $Z$ that contains sufficient information to predict the target $ Y$ while minimizing the information $ Z$ retains about $ X$, thereby enhancing generalization. 
Here, $ X$, $ Y$, and $ Z$ form a Markov chain where $ Z \leftrightarrow X \leftrightarrow Y$ \cite{vib2017}. 
We incorporate the additional modality information $ T $ and $ Y $ as conditions, assuming that $ T $ exhibits the same Markov property as $ X $, i.e., $Z \leftrightarrow T \leftrightarrow Y$. 
Introducing $ T $ as a condition allows the model to learn representations $ Z $ that incorporate beneficial information about $ Y $ from $ T $, thereby enhancing classification performance.  
In \cref{fig:com_mmib}, we provide a detailed comparison of our method with the information bottleneck and the conditional information bottleneck.
Thus, we obtain the overall constraint:
\begin{equation}
\label{Eq:target}
    \min I(Z;X\mid T, Y)-\beta I(Z;Y),
\end{equation}
where $\beta$ is the Lagrange multiplier and $I\left(\cdot;\cdot \right)$ denotes mutual information. Given \( x \in X \), \( y \in Y \), \( z \in Z \), and \( t \in T \), we derive the variational upper bound for Eq. \cref{Eq:target}:
\begin{equation}
\label{Eq:secondForm}
\begin{aligned}
&\langle\frac{\log e\left(z\mid x,t \right)}{\log r\left( z\mid t,y\right)}\rangle-\beta \langle\log c \left( y\mid z\right)\rangle\\
&= \mathbb{KL}\left[e\left(z\mid x,t \right)\mid\mid r\left( z\mid t,y\right) \right]-\beta \langle\log c\left(y\mid z\right)\rangle,
\end{aligned}
\end{equation}
where $r\left(\cdot\right)$ is the variational approximation, and $e\left(\cdot \right)$ and $c\left(\cdot \right)$ represent the encoder and classifier in our model, respectively. For brevity, we use $ \langle \cdot \rangle $ to denote $\mathbb{E}_{p\left(x,y,t \right)p\left(z\mid x,t\right)}$\cite{cib2020}. The detailed derivation of \cref{Eq:secondForm} will be presented in \cref{sec:appendix_objective}.

\paragraph{Variational approximation.} Due to the computational intractability of theoretical constraints, we need to incorporate the conditional information of $T$ and $Y$ into the variational approximation  $r$. 
Specifically, we aim to retain the information in modality $ T $ that is useful for classifying $ Y $, while removing other irrelevant information in $ T $, i.e., eliminating the overlapping components of $ T $ across different classes. 
To accomplish this, we naturally introduce the Gram-Schmidt orthogonalization process. 
Gram-Schmidt orthogonalization removes the components of other vectors in the direction of the target vector, thereby achieving orthogonality, which aligns perfectly with our objective.
Given the labeled modality information $T = \{t_1, t_2,\dots,t_n\}$ with corresponding $Y = \{ y_1, y_2,\dots, y_n\}$, where $n$ is the number of categories, 
we illustrate this with $n=2$ and apply Gram-Schmidt orthogonalization to obtain orthogonal text vectors:
\begin{equation}
\label{Eq:schimit}
\begin{aligned}
\mu_{r1}& = t_1,\\
\mu_{r2}& = t_2 - \frac{\left( t_2, \mu_{r1 }\right)}{\left( \mu_{r1 }, \mu_{r1 }\right)}\mu_{r1 }.
\end{aligned}
\end{equation}
Here, we use the normal distribution for the approximation:
\begin{equation}
\label{Eq:distributionR}
\begin{aligned}
r(z\mid y,t) = \mathcal{N}\left(\mathbf{\mu_r}, \mathbf{I} \right),
\end{aligned}
\end{equation}
where $\mathbf{\mu_r} = \left(\mu_{r1},\mu_{r2} \right)^T$. Notably, in practice, we employ normalized $\mathbf{\mu_r}$ and reduce the dimension of the hidden layer to explicitly compress the representation of $z$, rather than imposing direct constraints as in prior work \cite{vib2017,infogcn2022}. 
\paragraph{Training loss.} For the first term in \cref{Eq:secondForm}, considering the practical numerical stability and computational complexity of the KL divergence, we approximate it using Maximum-Mean Discrepancy (MMD) \cite{mmd2015}, which has been effectively employed in Zhao et al.'s work \cite{infovae2017} and Chi et al.'s work \cite{infogcn2022}. Following Chi et al. \cite{infogcn2022}, we simply define $\mathcal{L}_{\text{MMD}}$ as:
\begin{equation}
\label{Eq:finalForm}
\mathcal{L}_{\text{MMD}} = \mid\mid\mu_{e\left(z\mid x,t \right)} - \mu_{r\left( z\mid t,y\right)} \mid\mid ^2_2,
\end{equation}
where $\mu_e\left(z\mid x,t \right) = \frac{1}{\mid \mathcal{D}_y\mid}\sum_{(x_i, t_i, y_i)\in \mathcal{D}_y}\mathbb{E}_{e\left(z\mid x, t\right)}\left[z \right]$, and $\mathcal{D}_y = \{x_i, y_i\mid y_i = y\}$.
The second term is proportional to the negative cross-entropy loss, with detailed derivations in \cref{sec:appendix_objective}.
Consequently, our loss function $\mathcal{L}$ is defined as follows:
\begin{equation}
\label{Eq:totalLoss}
\mathcal{L} = \mathcal{L}_{\text{MMD}} + \beta\mathcal{L}_{cls},
\end{equation}
where $\beta$ is the Lagrange multiplier mentioned earlier in \cref{Eq:target}, which we treat as a hyperparameter.

\subsection{InfoFD}
\label{infofd}
In this section, we detail our proposed AI-generated image detection method. 
First, to address feature redundancy and leverage CLIP's multimodal capability, we propose the Text-Guided Conditional Information Bottleneck (TGCIB) method.
Then, In \cref{sec:preliminaries}, we found that random pooling text exhibits ``bias''.
To leverage this and the text's semantic content, we introduce Dynamic Text Orthogonalization (DTO), which dynamically weights and averages the text features in each batch with historical text features during training.
These modules are elaborated in the following subsections.

\paragraph{Text-Guided Conditional Information Bottleneck (TGCIB).} 
As highlighted in the work of Achille and Soatto \cite{emergence2018}, introducing perturbations to the input $ X $ can enhance the generalization capability of the learned representation $ Z $. Consequently, we incorporate Composite Gaussian Perturbation (CGP) into the images $I$.
For the image feature $ I $, we first augment it by adding CGP to obtain $ \tilde{I} $. 
The specific calculation of CGP is as follows:
\begin{equation}
    \label{Eq:cgp}
    \begin{aligned}
    \tilde{I} = \frac{I - \mu_I}{\sigma_I} + \epsilon_g, \epsilon_g \sim \mathcal{N}\left(0, \Sigma_s\right) + \text{Uniform}\left(0,1\right),
    \end{aligned}
\end{equation}
where  $\mu_I$ and $\sigma_I$ are the mean and standard deviation of $I$, and $\Sigma_s$ is a hyperparameter used to control the intensity of the augmentation.
CGP is inspired by data augmentation techniques in domain generalization, aiming to simulate multi-domain data distributions using single-domain data, which is particularly suitable for generalizable AI-generated detection tasks.
We then employ the reparameterization trick \cite{vae2013} to generate $ \mathbf{z} $. 
Specifically, for $ \epsilon \sim \mathcal{N}(0, \mathbf{I}) $, the representation $ \mathbf{z} $ is computed as $ \mathbf{z} = \mathbf{\mu_e} + \Sigma_e \epsilon $, where $ \mathbf{\mu_e} $ and $ \Sigma_e $ are learnable parameters derived from $ \tilde{I} $. 
Next, we utilize a simple linear layer followed by a Sigmoid function as the classifier to derive the probability $ \hat{y} $. 
During training, we additionally leverage $ \mathbf{\mu_e} $ and $ \mathbf{\mu_r} $ to compute $ \mathcal{L}_{\text{MMD}} $.

\paragraph{Dynamic Text Orthogonalization (DTO).}
To optimize text feature utilization, we leverage both semantic information and the global ``bias'' in the text, as indicated by our preliminary findings. 
Here, text features are treated as the additional modality $ T $. 
Initially, we split the input text features into real and fake image text features based on image categories and apply a linear layer to obtain projected text features $\mathbf{t} = (t_r, t_f)^T$. 
To leverage both CLIP's inherent ``bias'' and prompt semantics, DTO is used to obtain weighted pooling text features $\mathbf{\bar{t}}$ instead of direct text features. 
DTO comprises Pooling and Fusion.
Pooling projects text features into a hidden space and sums features for each group. 
Fusion includes Batch Fusion and Orthogonalization. 
In Batch Fusion, the model uses Pooling's feature sums and performs unbiased fusion with the previous batch's DTO results, as in \cref{Eq:DTO}. 
For the $ j $-th batch, Batch Fusion is calculated as follows:

\begin{equation}
\label{Eq:DTO}
\begin{aligned}
\mathbf{\bar{t}}^j_* &= \frac{1}{L_p + B_*}\left(\mathbf{\bar{t}}^{\left(j-1\right)}_* \cdot L_p + \sum^{B_*}_{i=1}\mathbf{t_{*i}}\right),\\
\end{aligned}
\end{equation}

where $L_p$ is a hyperparameter controlling the scale of global pooling, $B$ is the batch size of inputs, $*$ represents the group corresponding to either real or fake. 
In the Orthogonalization stage, we use Gram-Schmidt orthogonalization for this process, with the detailed calculation described in \cref{Eq:schimit}. 
Benefiting from the fact that the orthogonalized vector set is equivalent to the original vector set, this operation does not result in loss of the information from $T$ that we aim to preserve.

\section{Experiments}
\label{sec:5}
We present our main experimental results in this chapter. For details on hyperparameters (\cref{sec:appendix_exp_detail}), additional benchmark (\cref{sec:appendix_exp_aigc}), additional metrics (\cref{sec:appendix_exp_metrics}), robustness experiments (\cref{sec:appendix_robustness}),additional ablation studies (\cref{sec:appendix_exp_aas}) and additional experiments on visualizations (\cref{sec:appendix_exp_dft}) and others, please refer to \cref{sec:appendix_exp}.

\subsection{Evaluation protocols}
Here, we present the primary evaluation settings. Additional metrics (AUROC, F1, TPR95, etc.) are reported in \cref{sec:appendix_exp_metrics}.
\textbf{Evaluation Protocol 1 (EP1).} Following \citet{genimage2024}, we train our method on GenImage/SD1.4 train set and evaluate GenImage test set. We retrained baselines on GenImage/SDV1.4 and evaluated it on the test set. For methods that only released inference code without training code, we conducted evaluations using their official inference implementations and pretrained weights (marked with \dag).
\textbf{Evaluation Protocol 2 (EP2).} Due to the rapid advancement of generative models, we conducted more extensive experiments across a broader range of generative models to comprehensively evaluate our method's effectiveness. Here, we employ GenImage's SD V1.4 as the training set and use CO-SPY \cite{cospy2025} for assessment. The prompts are generated by InterVL \cite{internvl2024}. 

\subsection{Comparison with baselines}
\paragraph{Comparison under EP1 settings.} Under the EP1 experimental settings, we retrained several proposed baselines with publicly available training codes on the GenImage/SDV1.4 \cite{genimage2024} dataset, with the results presented in \cref{tab:ep1_gen}.
In \cref{tab:ep1_gen}, we report the accuracy (threshold = 0.5) for images generated by each model as well as the overall average accuracy. Our method demonstrates significant superiority across all metrics. Specifically,
in the results of the DTO-based approach, for all categories of test images infoFD shows an accuracy higher than 90\%. Meanwhile, the most significant improvement is in the ADM subcategory, which is higher than the second 17.17\%. We attribute this to the significant ``bias'' phenomenon of ADM in layer11, please refer to the visualisation \cref{fig:layer_adm} in \cref{sec:appendix_exp_vis} for details. Despite being up to 2\% below SOTA on models from SD, Wukong, and GLIDE, our method exhibits excellent robustness, resulting in an overall average accuracy of 97.82\% for our method.
\paragraph{Comparison under EP2 settings.} To validate the generalization capability of our method against more advanced generative models, we conduct evaluations on the CO-SPY dataset, with results presented in \cref{tab:ep1_cospy}. Notably, since CO-SPY only provides publicly available fake images, we utilize MSCOCO as the authentic dataset for evaluation, with random resampling performed for each subset to ensure diversity. Our method achieved outstanding performance across all sub-datasets, almost surpassing the second-best approach in both individual sub-dataset accuracy and average accuracy. This is attributed to the robust generalization capabilities stemming from our identification of textual ``bias'' and the proposed conditional information bottleneck.

\begin{table*}[h]
  \vskip -0.1in
  \caption{\textbf{Accuracy on GenImage \cite{genimage2024}.} Train on GenImage/SDV1.4, and evaluation on GenImage test set. All results are reported as \textbf{Mean (STD)}, computed from 3 independent runs with different random seeds.  The baselines results presented in the table were obtained through retraining on GenImage/SDV1.4. The highest values in each column are highlighted in \textbf{bold} and the second-highest results are indicated with \underline{underlining}. \textbf{\dag}: the results are evaluated by their pretrained weights. }
  \label{tab:ep1_gen}
  \centering
  \resizebox{\columnwidth}{!}{\begin{tabular}{cccccccccc}
      \toprule
      Method                         & BigGAN                 & MidJourney             & SD V1.4               & SD V1.5      & VQDM         & Wukong                & GLIDE                 & ADM                   & Avg. (STD)             \\
      \midrule
      CNNSpot \cite{cnn2020}         & 49.90 (0.04)           & 52.29 (0.91)           & 68.30 (3.27)          & 68.33 (3.22) & 50.08 (0.01) & 58.54 (1.57)          & 51.37 (0.54)          & 50.29 (0.15)          & 56.14 (1.21)           \\
      FreqNet \cite{freqnet2024}     & 76.03 (6.84) & 72.33 (6.89) & 90.33 (11.20) & 90.20 (11.32) & 53.80 (3.14) & 87.53 (9.57) & 68.10 (2.78) & 49.33 (0.49) & 73.46 (3.80)            \\
      Fusing \cite{fusing2022}       & 49.87 (0.02)           & 67.70 (0.35)           & \textbf{99.88 (0.01)} & \textbf{99.80 (0.00)} & 49.88 (0.02) & \textbf{99.20 (0.03)} & \textbf{99.42 (0.00)} & 49.92 (0.02)          & 70.80 (0.05)           \\
      UniFD \cite{unifd2023}         & 78.07 (0.83)           & 74.47 (0.32)           & 81.38 (0.49)          & 81.67 (0.09) & 54.37 (0.18) & 73.72 (0.46)          & 75.60 (0.07)          & 56.08 (0.08)          & 71.92 (0.27)           \\
      NPR \cite{npr2024}             & 63.97 (7.43)           & 77.50 (3.93)           & 99.00 (0.14)          & 98.87 (0.17) & 55.33 (1.08) & 95.90 (1.30)          & 88.53 (2.08)          & 63.87 (2.18)          & 80.37 (2.02)           \\
      DE-FAKE \cite{defake2023}           & 60.34 (0.34)                  & \underline{80.72 (6.78)}        & 87.59 (1.30)                & 83.30 (2.73)       & 67.22 (0.55)       & 77.91 (1.52)            & 84.55 (2.95)           & 60.42 (0.43)                  & 75.25 (0.99)                 \\
      FatFormer$^\dag$ \cite{fatformer2024}             & 96.79 (--)                & 56.08 (--)                  & 67.79 (--)                 & 68.04 (--)        & 86.85 (--)        & 73.04 (--)                  & 87.99 (--)                 & 78.42 (--)                & 76.88 (--)                  \\
      AIDE$^\dag$ \cite{aide2024}    & 66.89 (--)             & 79.38 (--) & 99.74 (--)            & \underline{99.76 (--)}   & 80.26 (--)   & 98.65 (--)            & 91.82 (--)            & 78.54 (--)            & 86.88 (--) \\
      C2P-CLIP$^\dag$ \cite{c2p2025} & \underline{97.94 (--)} & 59.12 (--)             & 82.76 (--)            & 82.73 (--)   & 87.16 (--)   & 80.12 (--)            & 88.48 (--)            & 77.12 (--)            & 81.93 (--)             \\
      \midrule
      Ours (Random)                  & 95.35 (1.14)           & 87.39 (1.26)           & \underline{99.81 (0.02)}          & 99.59 (0.02) & \textbf{94.89 (1.21)} & 98.67 (0.27)          & 97.77 (0.64)          & \underline{90.66 (0.02) }          & \underline{95.52 (0.72)}           \\
      Ours (DTO)                     & \textbf{98.55 (0.24)}  & \textbf{94.32 (1.54)}  & 98.71 (0.46)          & 98.52 (0.39) & \underline{94.79 (0.37)} & \underline{98.72 (0.39)}          & \underline{98.34 (0.11)}          & \textbf{96.59 (0.80)} & \textbf{97.82 (0.15)}  \\
      \bottomrule
    \end{tabular}}
  \vskip -0.1in
\end{table*}

\begin{table*}[h]
  \caption{\textbf{Accuracy on CO-SPY \cite{cospy2025}.} Train on GenImage \cite{genimage2024} /SDV1.4, and evaluate on CO-SPY. All results are reported as \textbf{Mean (STD)}, computed from 3 independent runs with different random seeds. The baseline results presented in the table were obtained through retraining on GenImage/SDV1.4. The highest values in each column are highlighted in \textbf{bold} and the second-highest results are indicated with \underline{underlining}. \textbf{\dag}: the results are evaluated by their pretrained weights.} 
  \label{tab:ep1_cospy}
  \centering
  \resizebox{\columnwidth}{!}{\begin{tabular}{@{}cccccccccc@{}}
  \toprule
                            & CNNSpot \cite{cnn2020} & FreqNet \cite{freqnet2024} & Fusing \cite{fusing2022} & NPR \cite{npr2024} & DE-FAKE \cite{defake2023} & FatFormer$^\dag$ \cite{fatformer2024} & AIDE$^\dag$ \cite{aide2024} & C2P-CLIP$^\dag$ \cite{c2p2025} & Ours         \\
  \midrule
  LDM                       & 74.66 (0.91) & 74.63 (1.87) & 90.83 (0.47) & 88.67 (2.90) & 70.90 (2.15) & 86.22 (--) & \underline{94.43 (--)} & 92.47 (--) & \textbf{99.47 (0.11)} \\
  SD V1.4                   & 73.29 (1.64) & 92.30 (7.22) & 98.43 (0.15) & 97.03 (2.15) & 75.97 (0.98) & 56.86 (--) & \textbf{99.57 (--)} & 77.21 (--) & \underline{99.14 (0.05)} \\
  SD V1.5                   & 73.37 (1.79) & 92.40 (7.29) & 98.44 (0.12) & 97.27 (2.03) & 75.15 (1.40) & 56.34 (--) & \textbf{99.51 (--)} & 77.07 (--) & \underline{99.45 (0.12)} \\
  SSD-1B                    & 66.82 (0.05) & 39.60 (7.16) & 39.21 (0.08) & 36.80 (4.78) & 74.60 (0.85) & 55.54 (--) & \underline{68.05 (--)} & 61.30 (--) & \textbf{99.20 (0.06)} \\
  Tiny-SD                   & 67.62 (0.05) & 85.97 (7.48) & \underline{93.55 (0.93)} & 91.70 (2.13) & 75.78 (2.02) & 70.94 (--) & 92.82 (--) & 85.11 (--) & \textbf{99.46 (0.12)} \\
  SegMoE-SD                 & 69.34 (0.10) & 88.27 (6.21) & 95.15 (0.06) & 96.97 (1.06) & 71.53 (5.27) & 43.49 (--) & \underline{97.98 (--)} & 52.71 (--) & \textbf{99.48 (0.14)} \\
  Small-SD                  & 68.03 (0.12) & 85.60 (8.15) & 93.93 (1.26) & 89.33 (3.72) & 72.53 (1.27) & 64.35 (--) & \underline{94.52 (--)} & 85.95 (--) & \textbf{99.42 (0.12)} \\
  SD V2.1                   & 68.90 (0.41) & 75.63 (10.55) & \underline{91.76 (0.01)} & 66.60 (9.68) & 75.95 (0.30) & 44.52 (--) & 86.25 (--) & 52.18 (--) & \textbf{98.40 (0.27)} \\
  SD V3                     & 71.87 (1.04) & \underline{77.43 (6.58)} & 67.95 (2.52) & 77.03 (8.17) & 77.12 (4.12) & 44.67 (--) & 73.64 (--) & 44.81 (--) & \textbf{99.29 (0.02)} \\
  SDXL-turbo                & 73.34 (1.50) & 63.60 (6.18) & 65.99 (5.35) & 47.57 (12.19) & 69.62 (4.62) & 56.85 (--) & \underline{79.33 (--)} & 56.94 (--) & \textbf{99.45 (0.12)} \\
  SD V2                     & 67.86 (0.38) & 78.10 (9.51) & \underline{86.42 (0.61)} & 71.27 (6.86) & 75.83 (0.83) & 56.61 (--) & 82.32 (--) & 56.73 (--) & \textbf{99.14 (0.05)} \\
  SD XL                     & 66.72 (0.01) & 42.90 (12.16) & 30.59 (0.02) & 32.90 (2.37) & \underline{76.83 (0.67)} & 64.91 (--) & 59.72 (--) & 54.92 (--) & \textbf{99.40 (0.13)} \\
  PG V2.5-1024              & \underline{66.78 (0.02)} & 42.53 (12.71) & 37.26 (2.63) & 39.30 (7.99) & 58.75 (5.12) & 51.96 (--) & 61.48 (--) & 45.08 (--) & \textbf{99.28 (0.03)} \\
  PG V2-1024                & 67.06 (0.02) & 79.93 (5.60) & 80.47 (0.00) & 77.23 (13.05) & 69.90 (1.35) & 64.48 (--) & \underline{86.19 (--)} & 43.58 (--) & \textbf{99.34 (0.04)} \\
  PG V2-512                 & 68.64 (0.29) & 84.67 (6.04) & 76.72 (2.79) & 83.27 (4.69) & 76.25 (1.25) & 58.87 (--) & \underline{83.99 (--)} & 68.99 (--) & \textbf{99.36 (0.03)} \\
  PG V2-256                 & 68.59 (0.19) & \underline{81.23 (5.90)} & 70.49 (0.78) & 56.60 (8.03) & 67.85 (2.15) & 52.95 (--) & 75.22 (--) & 51.05 (--) & \textbf{98.56 (0.33)} \\
  PAXL V2-1024              & 67.67 (0.02) & 74.17 (3.43) & 72.72 (1.40) & 76.50 (10.66) & 70.10 (1.35) & 59.73 (--) & \underline{77.23 (--)} & 40.78 (--) & \textbf{99.07 (0.09)} \\
  PAXL V2-512               & 69.20 (0.04) & 80.20 (3.36) & 75.49 (0.64) & 83.80 (6.85) & 68.70 (2.45) & 57.25 (--) & \underline{90.52 (--)} & 44.07 (--) & \textbf{99.44 (0.09)} \\
  LCM-SD XL                 & 71.41 (0.98) & 65.23 (7.86) & 46.85 (2.72) & 77.97 (17.15) & \underline{78.95 (2.70)} & 52.49 (--) & 53.57 (--) & 63.24 (--) & \textbf{99.46 (0.11)} \\
  LCM-SD V1.5               & 77.10 (1.92) & 85.30 (5.47) & 76.24 (0.59) & \underline{90.67 (4.88)} & 78.62 (0.12) & 38.97 (--) & 87.43 (--) & 49.06 (--) & \textbf{99.32 (0.02)} \\
  FLUX.1-sch                & 68.23 (0.04) & \underline{77.00 (5.14)} & 37.53 (1.79) & 73.57 (10.42) & 71.80 (1.95) & 34.48 (--) & 65.68 (--) & 37.02 (--) & \textbf{93.21 (0.81)} \\
  FLUX.1-dev                & 69.58 (0.36) & 74.70 (3.92) & 49.49 (0.59) & \underline{80.83 (7.35)} & 66.03 (2.27) & 34.61 (--) & 79.69 (--) & 33.98 (--) & \textbf{97.54 (0.25)} \\
  \midrule
  Avg.                      & 70.64 (0.51) & 74.60 (3.45) & 71.61 (0.31) & 74.23 (6.46) & 72.67 (4.68) & 54.87 (--) & \underline{81.32 (--)} & 57.92 (--) & \textbf{98.92 (0.02)} \\
  \bottomrule
  \end{tabular}}
  \vskip -0.1in
  \end{table*}

\subsection{Ablation Studies}
\begin{wraptable}{r}{0.4\columnwidth}
  \vskip -0.18in
	\caption{\textbf{Ablation Study.} All results are reported as \textbf{Mean (STD)}, computed from 3 independent runs with different random seeds.}
	\label{tab:main_ablation}
	\centering
  \resizebox{0.39\columnwidth}{!}{
  \begin{tabular}{ccccccl}
		\toprule
		\multicolumn{2}{c}{Components}  & \multicolumn{3}{c}{Conditions} &      \multirow{2}{*}{Acc(\%) } \\
		\cmidrule(lr){1-2}  \cmidrule(lr){3-5}     
		CGP & $ \mathcal{L}_{\text{MMD}} $     & $\mathcal{N}(\mathbf{0},\mathbf{1})$ & $Y$               & $T$  & \\
		\midrule
		           &            &            &            &            & 91.57 (2.50)  \\
		\checkmark &            &            &            &            & 86.42 (1.33)  \\
		           & \checkmark & \checkmark &            &            & 94.26 (2.73) \\
		\checkmark & \checkmark & \checkmark &            &            & 84.65 (2.84)  \\
		\checkmark & \checkmark & \checkmark & \checkmark &            & 93.50 (2.53)  \\
		\checkmark & \checkmark &            &            & \checkmark & 72.16 (2.70)  \\
		\checkmark & \checkmark &            & \checkmark & \checkmark & \textbf{97.82 (0.15)}  \\
		\bottomrule
	\end{tabular}}
  \vskip -0.1in
\end{wraptable}
We conducted ablation studies on InfoFD's modules and different conditions, with results presented in \cref{tab:main_ablation}. Notably, removing both CGP and $ \mathcal{L}_{\text{MMD}} $ reduces InfoFD to essentially CLIP with a simple reparameterization module.  
The combined use of CGP with conditional information $ Y $ enhances overall model performance, whereas employing CGP alone leads to significant performance degradation. We attribute this phenomenon to the randomness introduced by CGP's noise. When the hidden layer $ Z $ lacks class-discriminative information (i.e., without $ Y $), the noise added by CGP obscures the distinction between real and fake features, resulting in misclassification and impaired performance. Conversely, when incorporating $ Y $'s information, despite the same noise level, the model can effectively identify feature classes, thereby improving generalization capability.  
Moreover, introducing either $\mathcal{N}(\mathbf{0},\mathbf{1})$ or $T$ alone as conditions also leads to performance degradation, with the latter exhibiting more severe deterioration. The performance drop induced by $\mathcal{N}(\mathbf{0},\mathbf{1})$  stems from the optimization conflict in information bottleneck compression, as analyzed in \cref{sec:1}. The degradation caused by $T$ can be attributed to two factors: (1) ambiguous categorical text features blur the decision boundaries between real and fake features in $Z$, and (2) the inherent modality gap hinders model convergence.
    


\begin{figure}
  \centering
  \begin{subfigure}[t]{0.66\linewidth}
    \includegraphics[width=\linewidth]{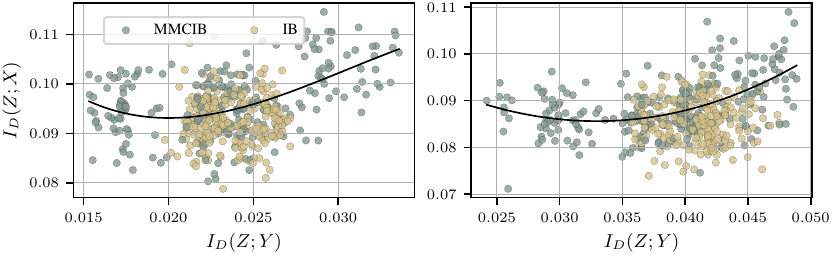}
    \vskip -0.06in
    \caption{}
    \label{fig:izx_izy}
  \end{subfigure}
  \begin{subfigure}[t]{0.33\linewidth}
    \includegraphics[width=\linewidth]{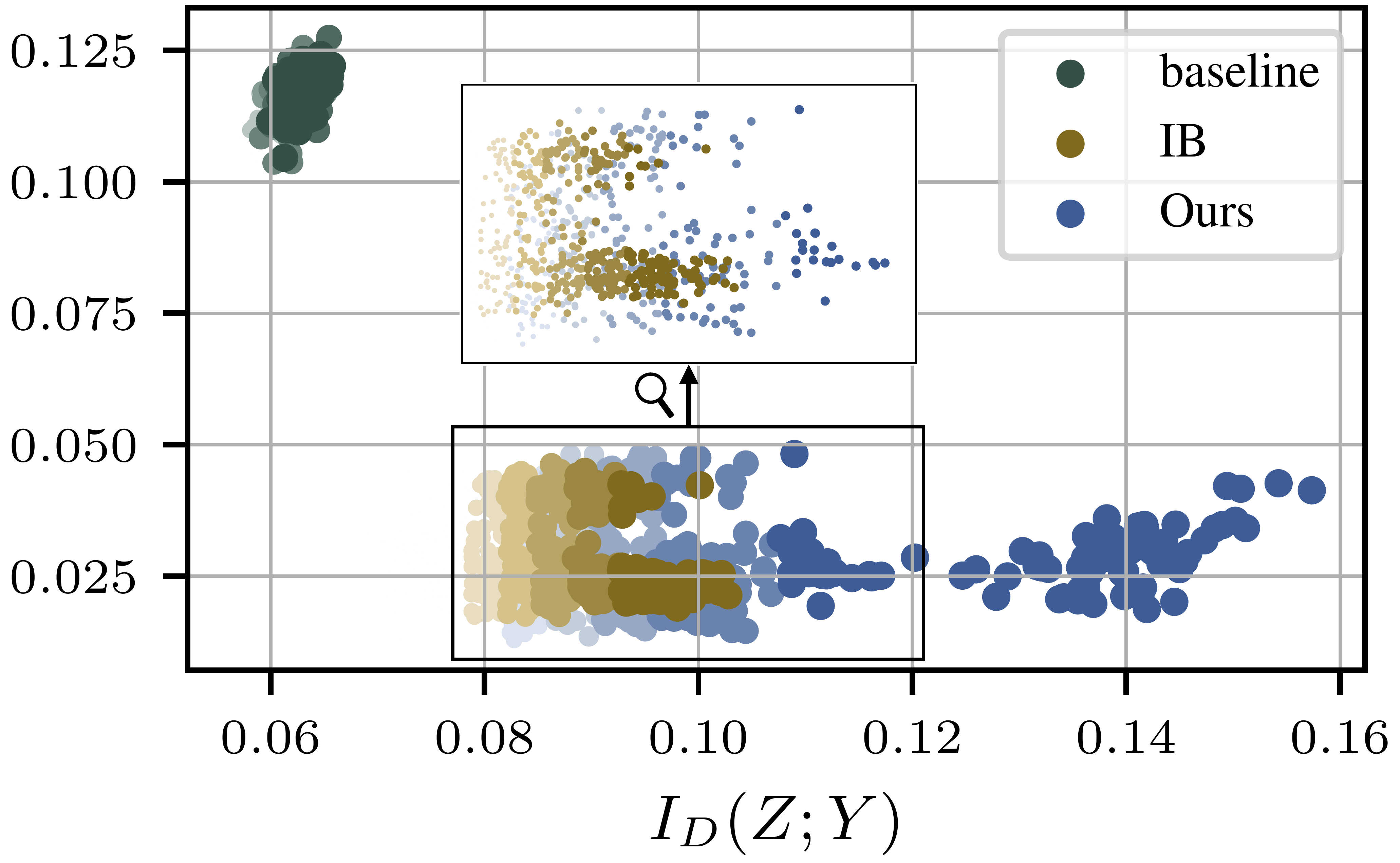}
    \vskip -0.06in
    \caption{}
    \label{fig:feature_redundancy}
  \end{subfigure}
    \vskip -0.05in
    \caption{\textbf{(a)} $I_D(X;Z)$ against $I_D(Z;Y)$ on MMCIB and IB. The \textbf{black line} represents the trend of MMCIB obtained through triple curve fitting. The results are calculated on GenImage \cite{genimage2024} Test set.
    \textbf{(b)} Illustration of feature redundancy. The darker the color, the greater the degree of compression. The baseline is controlled using different sizes of hidden dimensions. }
    \vskip -0.22in
  \end{figure}

\paragraph{Mutual Information Estimation.}
\label{sec:5_5_mie}
We present a comparative visualization of mutual information metrics between our method and standard information bottleneck approaches, specifically analyzing $ I(Z;Y) $ and $ I(Z;X) $. Given the challenges in high-dimensional mutual information estimation \cite{mine2018,diffmi2024}, we employ the \textbf{D}iffusion \textbf{S}pectral \textbf{E}ntropy (\textbf{DSE}) proposed in \cite{diffmi2024} to quantify $ {I}_D(Z;X) $ and $I_D(Z;Y) $. We briefly introduce this work in \cref{sec:appendix_mie}. For each epoch, we randomly select 1,000 data points from both GAN-generated and diffusion-generated images, repeating this process 7 times to ensure statistical reliability. Our method demonstrates significant advantages. As shown in \cref{fig:izx_izy}, the conditional information bottleneck method preserving the information in $X$ that is relevant to $Y$ during the compression process. The high correlation between MMCIB's $ I_D(Z;Y) $ and $ I_D(Z;X) $ indicates that our method preserves $ Y $-relevant information in $ X $ and compresses it into $ Z $. Additionally, compared to the standard information bottleneck method, our method achieves higher $ I_D(Z;Y) $, demonstrating the advantage of introducing $ T $ as additional information. Please refer to \cref{sec:appendix_mie} for more estimation results.
We utilized a standard linear layer, IB, and our method to visualize mutual information estimation, recording the mutual information between $I_D(Z;X)$ and $I_D(Z;Y)$ under varying degrees of feature compression, resulting in \cref{fig:feature_redundancy}. \cref{fig:feature_redundancy} demonstrates the preservation capability of our conditional information bottleneck method for $Y$-related information, achieving higher $I_D(Z;Y)$ and lower $I_D(Z;X)$ compared to the baseline and IB. Additionally, \cref{fig:feature_redundancy} confirms the existence of feature redundancy.
\paragraph{PCA 2D Projection.}

\begin{figure}[ht]
  \vskip -0.2in
  \centering
  \begin{subfigure}[t]{0.23\columnwidth}
      \centering
      \includegraphics[width=\columnwidth]{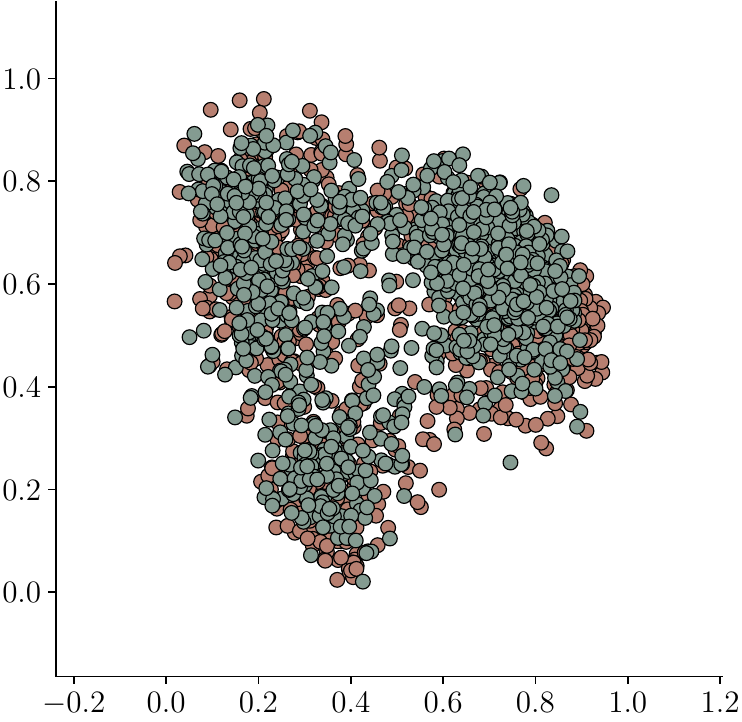}
      \caption{ CLIP - ADM.}
      \label{fig:adm_clip_vis}
  \end{subfigure}
  \begin{subfigure}[t]{0.23\columnwidth}
      \centering
      \includegraphics[width=\columnwidth]{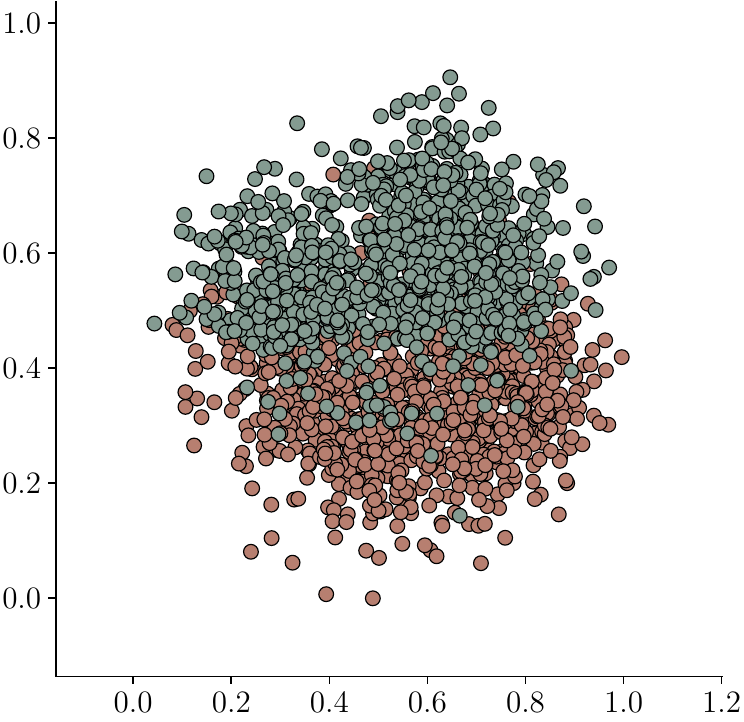}
      \caption{ $z$ - ADM.}
      \label{fig:adm_z_ty_vis}
  \end{subfigure}
  \centering
  \begin{subfigure}[t]{0.23\columnwidth}
      \centering
      \includegraphics[width=\columnwidth]{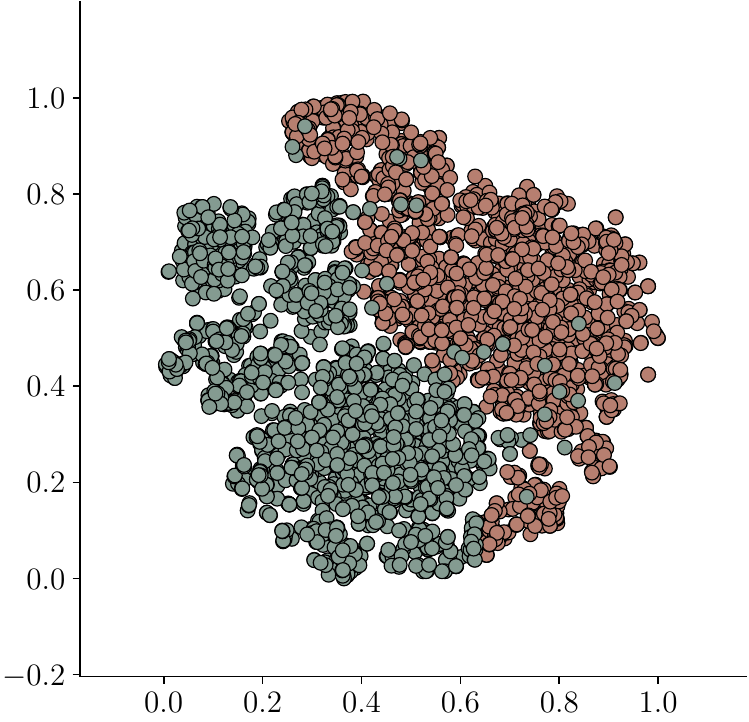}
      \caption{ CLIP - BigGAN.}
      \label{fig:biggan_clip_vis}
  \end{subfigure}
  \begin{subfigure}[t]{0.23\columnwidth}
      \centering
      \includegraphics[width=\columnwidth]{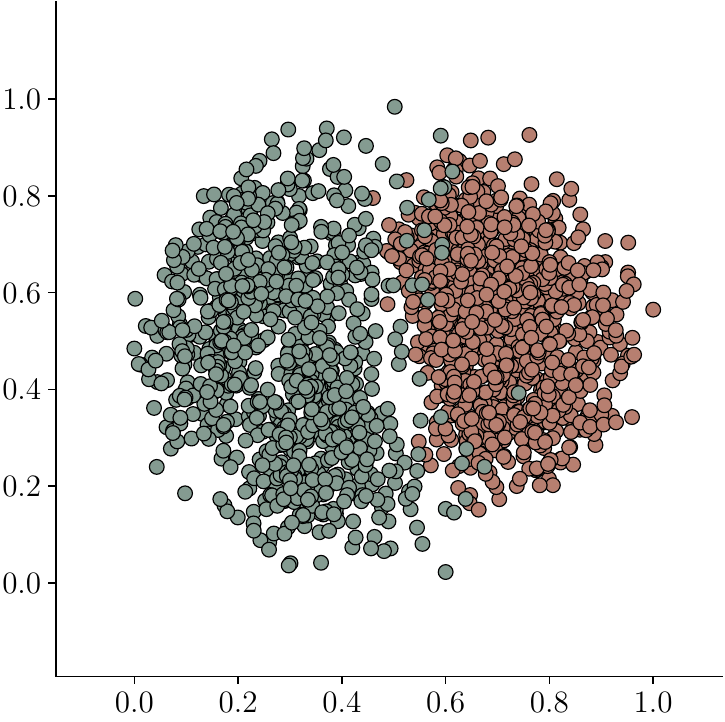}
      \caption{ $z$ - BigGAN.}
      \label{fig:biggan_z_ty_vis}
  \end{subfigure}
  \vskip -0.05in
  \caption{\textbf{PCA Projection.} \textcolor{green}{Green}: real image features; \textcolor{red}{Red}: fake image features; CLIP: CLIP feature projections; $z$: TGIB-extracted representations.}
  \label{fig:pca_visualization}
\vskip -0.1in
\end{figure}
We conducted PCA on the features extracted by the ViT and projected them into a 2D space, with the results shown in \cref{fig:pca_visualization}.
As shown in \cref{fig:pca_visualization}, the representations $ z $ learned by the model exhibit a more pronounced distinction between real and fake images compared to CLIP's original features. 
The intra-class distances are more compact, and the distributions approximate a Gaussian form, which further facilitates the final classifier's discriminative performance.

\section{Conclusion}
\label{sec:6}

In this paper, we propose a generalisable multimodal conditional information bottleneck method to mitigate feature redundancy in CLIP, with the goal of enhancing its performance by incorporating modality-specific information during training. Additionally, we examine the impact of text features on improving CLIP’s ability to detect AI-generated images. Extensive ablation experiments validate our findings. 
In future work, we aim to (i) explore CLIP’s feature space to uncover intrinsic characteristics and (ii) refine the multimodal conditional information bottleneck framework to further optimise performance.


\bibliography{refs.bib}
\bibliographystyle{unsrtnat}
\appendix
\newpage
\section{Additional Discussion}
\subsection{Impact Statement}
\label{sec:appendix_impact}
Our work aims to advance the field of machine learning by investigating the relationship between CLIP's text features and real versus fake images, as well as proposing a multimodal conditional information bottleneck method. 
This research carries significant potential societal impacts. 
On the positive side, our method can mitigate the spread of rumors fueled by forged images, reduce misinformation, and enhance the credibility of media. 
Conversely, our findings on the relationship between CLIP's text features and real versus fake images could potentially be exploited for malicious purposes. 
We urge the research community to harness its positive implications.
\subsection{Limitations}
\label{sec:appendix_limitations}
Although we have identified the ``bias'' phenomenon and proposed infoFD to address both this issue and feature redundancy, demonstrating excellent performance, certain limitations remain. First, the underlying causes of the bias remain unclear. While this presents significant challenges, we are committed to investigating it in future work. Additionally, constrained by workload and other factors, we have not conducted in-depth research or experiments on the specific application effects of the multimodal conditional information bottleneck across different modalities or domains.
\section{Supplementary on Information Bottleneck}
\subsection{Detailed Derivation of the Formula}
\label{sec:appendix_objective}
\cref{Eq:target} can be decomposed into:
\begin{equation}
    \label{Eq:furtherDecomposed}
    \begin{aligned}
        \min  I(Z;X\mid Y,T)-\beta I(Z;Y) &= H\left(Z\mid Y,T \right) - H\left(Z\mid X, Y,T \right)-\beta I(Z;Y)\\
        &=\min <\log e\left(z\mid x,t \right)> - <\log p\left( z\mid t,y\right)>-\beta I(Z;Y).
        \end{aligned}
\end{equation}
As in prior work  \cite{vib2017,emergence2018,cib2020}, due to the intractability of $ p(z \mid t, y) $ in practice, we use $ r(z \mid t, y) $ as a variational approximation to $ p(z \mid t, y $):
\begin{equation}
    \small
    \label{Eq:variationalApproximation}
    \begin{aligned}
        &<\log e\left(z\mid x,t \right)> - <\log p\left( z\mid t,y\right)>-\beta I(Z;Y)\\
        &= <\log e\left(z\mid x,t \right)> -<\log r\left( z\mid t,y\right)> + <\log r\left( z\mid t,y\right)>-<\log p\left( z\mid t,y\right)>-\beta I(Z;Y)\\
        &= <\log e\left(z\mid x,t \right)> -<\log r\left( z\mid t,y\right)> - \mathbb{KL}\left(p\mid\mid r\right)-\beta I(Z;Y)\\
        &\leq  <\log e\left(z\mid x,t \right)> -<\log r\left( z\mid t,y\right)>-\beta I(Z;Y)\\
        &= <\frac{\log e\left(z\mid x,t \right)}{\log r\left( z\mid t,y\right)}>-\beta I(Z;Y),
        \end{aligned}
\end{equation}
where $ \mathbb{KL}\left( \cdot \mid \mid \cdot \right) $ denotes the Kullback-Leibler divergence between the two distributions.
Next, we first address the treatment of the last term, which can be transformed as follows:
\begin{equation}
    \label{Eq:secondTerm}
    \begin{aligned}
        -\beta I(Z;Y) &= -\beta\left[H\left(Y \right) - H\left(Y\mid Z\right) \right]\\
        &= -\beta H\left(Y\right) + \beta H\left(Y\mid Z\right),
        \end{aligned}
\end{equation}
since $H(Y)$ is a constant, we omit it and rewrite the second term as:
\begin{equation}
    \label{Eq:propto}
    \begin{aligned}
        -\beta I(Z;Y) &\propto \beta H\left(Y\mid Z\right)\approx\beta\frac{1}{\mid \mathcal{D}_y\mid}\sum_{\left(x_i y_i\right)\in \mathcal{D}_y}\mathbb{E}_{z\sim e\left(z|x_i\right)}[-\log p\left(y_i|z\right)].
        \end{aligned}
\end{equation}
where $\mathcal{D}_y = \{x_i, y_i\mid y_i = y\}$.
Noting that the rightmost term in \cref{Eq:propto} simplifies to the cross-entropy loss function, we denote it as $ \mathcal{L}_{cls} $.

Next, we discuss the simplification of the first term in \cref{Eq:variationalApproximation}. We replace the first term with an approximation based on the Maximum Mean Discrepancy (MMD) \cite{mmd2015,infogcn2022}:
\begin{equation}
    \label{Eq:firstTerm}
    \begin{aligned}
         &<\frac{\log e\left(z\mid x,t \right)}{\log r\left( z\mid t,y\right)}>\\
        &=  \mathbb{KL}\left[e\left(z\mid x,t \right)\mid\mid r\left( z\mid t,y\right) \right]\\
        &\approx \mid\mid\mathbb{E}\left[\phi\left(e\right)\right]-\mathbb{E}\left[\phi\left(r\right)\right]\mid\mid_{\mathcal{H}}^{2},\\
    \end{aligned}
\end{equation}
where $\mathcal{H}$ is the Reproducing Kernel Hilbert Space (RKHS) and $\phi$ is the mapping function.
To simplify the computation, we define $\phi$ as the identity mapping and set $\mathcal{H} = \mathbb{R}^d$, where $d$ is the dimension of the hidden state, resulting in the final simplified form:
\begin{equation}
    \label{Eq:finalFirstTerm}
    \begin{aligned}
         &<\frac{\log e\left(z\mid x,t \right)}{\log r\left( z\mid t,y\right)}>\\
        &\approx \mid\mid\mu_{e\left(z\mid x,t \right)} - \mu_{r\left( z \mid t,y\right)} \mid\mid ^2_2\\
        &=\mathcal{L}_{\text{MMD}}.
    \end{aligned}
\end{equation}
Ultimately, we obtain the total loss function:
\begin{equation}
    \label{Eq:finalLoss}
    \mathcal{L} = \mathcal{L}_{\text{MMD}} + \beta \mathcal{L}_{cls}.
\end{equation}

\subsection{Mutual Information Estimation }
\label{sec:appendix_mie}
While the information bottleneck method provides an elegant theoretical framework for modeling neural networks through mutual information \cite{vib2017,cib2020,benchmi2023}, its practical implementation encounters significant challenges due to the curse of dimensionality when processing high-dimensional data \cite{mine2018,infomax2019,benchmi2023}. The exponential growth in the number of bins required for probability histogram estimation with increasing dimensions makes mutual information computation particularly difficult \cite{mine2018}. To address this, numerous neural network-based mutual information estimators have been developed. In this work, we employ Diffusion Spectral Entropy \cite{diffmi2024}, which circumvents the binning process through manifold learning techniques, yielding an approximate equivalent substitute for mutual information denoted as $I_D(\cdot;\cdot)$. \citet{diffmi2024} provided a theoretical and empirical validation of $I_D(\cdot;\cdot)$'s effectiveness, robustness, and practical utility in their work. For a detailed technical exposition of Diffusion Spectral Entropy (DSE), we refer readers to their original paper.
\section{Supplementary on Preliminaries}
\label{sec:appendix_exp_preliminaries}
\subsection{Broader Experiments about ``bias''}

\begin{figure}[ht]
     \includegraphics[width=\columnwidth]{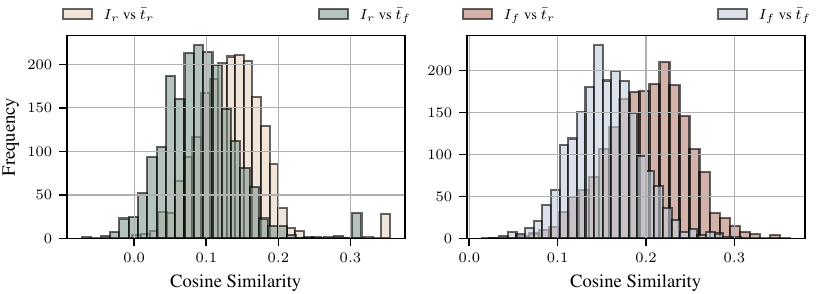}
    \caption{\textbf{Frequency histogram of cosine similarity between text features and image features.} The left figure shows the frequency plot of cosine similarity between real images and $\bar{t}_r$ and $\bar{t}_f$, while the right figure displays the frequency plot of cosine similarity between fake images and $\bar{t}_r$ and $\bar{t}_f$. We randomly sampled 2000 fake and real images.}
    \label{fig:real_fake_images_hist_on_train}
 \end{figure}

\paragraph{Between Corresponding Prompts.}
A natural question arises from the preliminaries' findings: if pooling text features without semantic information exhibit higher similarity with fake versus real images, how do semantically meaningful text features behave in this context? To Answer this, we conduct the analysis experiment on LAION \cite{laion2021} and DiffusionDB \cite{diffusiondb2022} datasets same as preliminaries. Specifically, we use LAION's images and corresponding as real images and pooling prompts, denoted as $I_r$ and $\bar{t_r}$. Similarly, we denote DiffusionDB' images as $I_f$ and pooling corresponding prompts as $\bar{t_f}$. The results for $ \cos(I_r, \bar{t}_r) $, $ \cos(I_r, \bar{t}_f) $, $ \cos(I_f, \bar{t}_r) $, and $ \cos(I_f, \bar{t}_f) $ are shown in \cref{fig:real_fake_images_hist_on_train}.

\begin{figure}[ht]
    \includegraphics[width=\columnwidth]{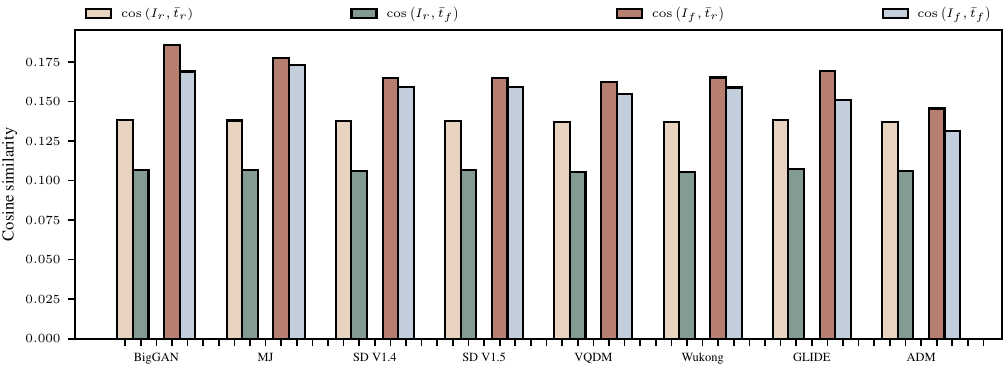}
    \caption{\textbf{Mean cosine similarity of different datasets with the same pooling real and fake text features.} A total of 1000 real images and 1000 fake images were used, with the fake images sourced from the corresponding generative models in the GenImage \cite{genimage2024} dataset. The real images were obtained from ImageNet.}
    \label{fig:diff_datasets_pooling_prompts}
\end{figure}

\cref{fig:real_fake_images_hist_on_train} illustrates that pooling text features with semantic information and their corresponding images exhibit a certain degree of distance in the CLIP feature space. Anomalously, the cosine similarity between $I_f$ and $\bar{t}_r$ remains slightly higher than that between $I_f$ and $\bar{t}_f$. One plausible hypothesis is that the pooling text features derived from real image prompts exhibit certain specific patterns that coincidentally align more closely with fake image features. To validate this conclusion, we preserved the pooling text features computed on LAION and DiffusionDB, and conducted additional experiments across multiple generative model datasets. The results are presented in the \cref{fig:diff_datasets_pooling_prompts}.

\cref{fig:diff_datasets_pooling_prompts} illustrates the difference in cosine similarity even in the case of mismatch between text features and image feature sources (or, rather, mismatch in semantic information). It is worth noting that for images generated by the generative model, $I_f$ and $\bar{t}_r$ still exhibit higher cosine similarity than $I_f$ and $\bar{t}_f$, which validates our previous conclusion that there is a specific pattern present in the textual features of the pooling real images that makes them more similar to the fake images. In addition, \cref{fig:diff_datasets_pooling_prompts} demonstrates that using mismatched text features still provides a certain degree of discriminative ability between real and fake images, which supports our conclusion regarding CLIP's ``bias''.

\begin{figure}[ht]
    \includegraphics[width=\columnwidth]{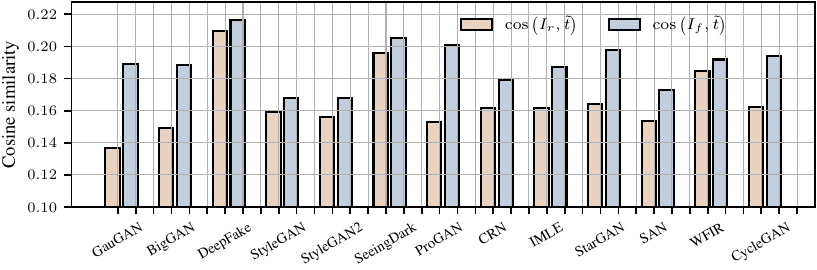}
    \caption{\textbf{Mean cosine similarity of different datasets with the same pooling real and fake text features on UniFD \cite{unifd2023}.}}
    \label{fig:unifd_bias}
\end{figure}

\begin{figure}[ht]
    \includegraphics[width=\columnwidth]{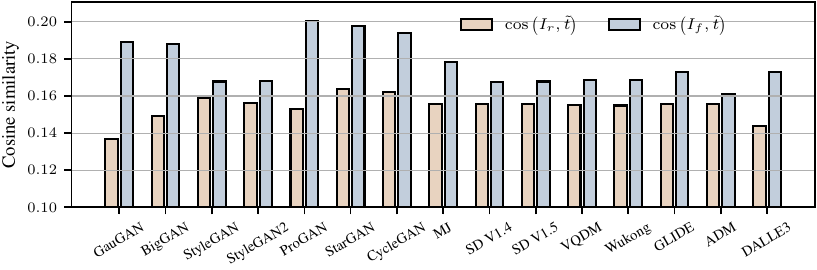}
    \caption{\textbf{Mean cosine similarity of different datasets with the same pooling real and fake text features on GANs and Diffusions.}}
    \label{fig:selected_bias}
\end{figure}

\paragraph{More Experiments.} We conduct same semantic analysis experiment on more GAN and Diffusion models, with results shown in \cref{fig:unifd_bias,fig:selected_bias}. 
\subsection{Layers Experiments}
\label{sec:appendix_layers}

\paragraph{Layer Diff.} To further analyze, we recorded the [CLS] outputs of different blocks in the ViT-L/14 encoder of CLIP, performed projection transformations, calculated the cosine similarity with $\bar{t}$, and took the mean, resulting in the findings shown in \cref{fig:layer_exp_in_pre}.

\cref{fig:layer_exp_in_pre} indicates that the output of the last layer of the ViT encoder in CLIP is not optimal. The ``bias'' exhibits a more pronounced discriminative ability in the output of the 12th block.
If we treat random text features as a neutral criterion for discrimination, then \cref{fig:layer_exp_in_pre} illustrates that not all layers of the ViT in CLIP contribute positively to AI-generated detection. 
Based on the differences in cosine similarity between real and fake images in \cref{fig:layer_exp_in_pre}, the features extracted by layers 1 to 12 of the ViT exhibit a beneficial effect for AI-generated detection. 
However, once past the 12th layer, the difference between real and fake image features diminishes significantly, even approaching zero,
and this difference remains at a low level even in the final layer.
Furthermore, our analysis reveals that the initial layers of the network predominantly capture low-level texture details, which play a crucial role in distinguishing discrepancies within manipulated images.

\begin{figure*}[ht]
        \centering
        \begin{subfigure}[t]{0.45\textwidth}
            \centering
            \includegraphics[width=\textwidth]{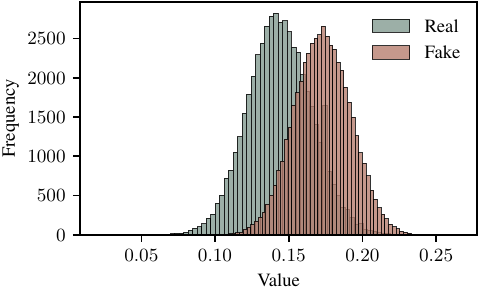}
            \caption{The layer23 (last) outputs.}
            \label{fig:train_layer23}
        \end{subfigure}
        \begin{subfigure}[t]{0.45\textwidth}
            \centering
            \includegraphics[width=\textwidth]{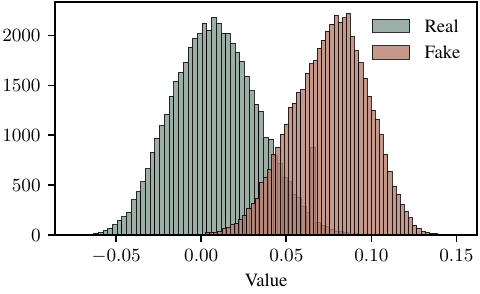}
            \caption{The layer11 outputs.}
            \label{fig:train_layer11}
        \end{subfigure}
        \caption{\textbf{Frequency histograms of cosine similarity between the outputs of different layers of the ViT in CLIP and random text features for the DiffusionDB and LAION datasets.} There are 50,000 real images and 50,000 fake images. All output image features were transformed using the projection layer into a 768-dimension feature space.}
        \label{fig:train_layers}
    \end{figure*}

Further more, we performed a statistical analysis of the similarity between the output features of various layers and random text features, yielding frequency histograms depicted in \cref{fig:train_layers}. 
The histograms reveal notable disparities among the output features of different layers, highlighted by the pronounced similarity gap between the features of layer 11 and random text features, as compared to that of layer 23. 
We posit that the features extracted by CLIP in the initial layers (e.g., layers 0 to 11) predominantly encapsulate fine-grained image details, which are particularly advantageous for AI-generated detection. 
Conversely,  as \citet{interpretclip2024} point out in their awesome work, the subsequent layers (e.g., layer 20 to 23) concentrate more on high-level semantic information, which offers limited utility in AI-generated detection. 
Despite the residual structures within these latter layers theoretically retaining semantic information from earlier layers, this information may progressively erode during intricate feature transformations (e.g., as a result of LayerNorm normalization processes).

\begin{figure}[ht]
    \centering
    \includegraphics[width=0.8\columnwidth]{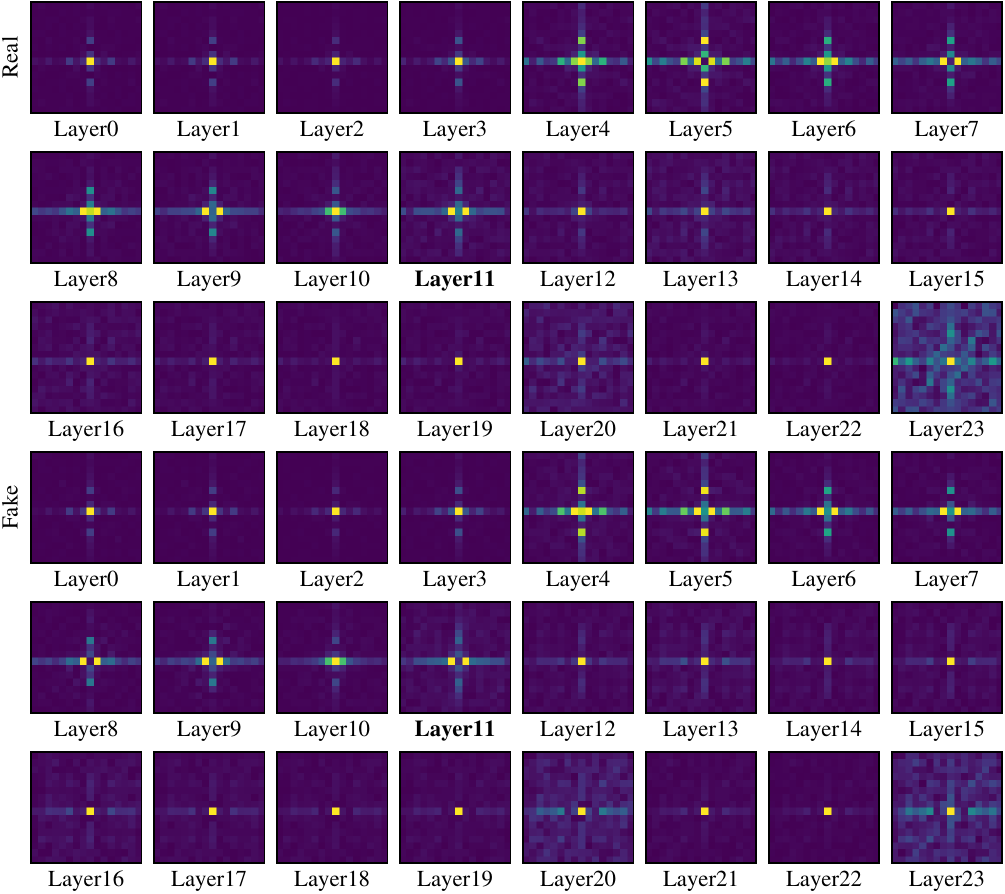}
    \caption{\textbf{DFT results of intermediate feature maps from Stable Diffusion.} We processed 200 fake images from Stable Diffusion and 200 real images from ImageNet, extracting intermediate layer outputs excluding [CLS] and reshaping them into 16x16 patches. Summing along channels produced 16x16 patch representations for real and fake images, followed by DFT and shift operations. Central regions represent low-frequency signals, while outer regions represent high-frequency signals.} 
    \label{fig:sd_layers_fft}
\end{figure}

\begin{figure}[ht]
    \centering
    \begin{subfigure}[t]{\columnwidth}
        \centering
        \includegraphics[height=2in]{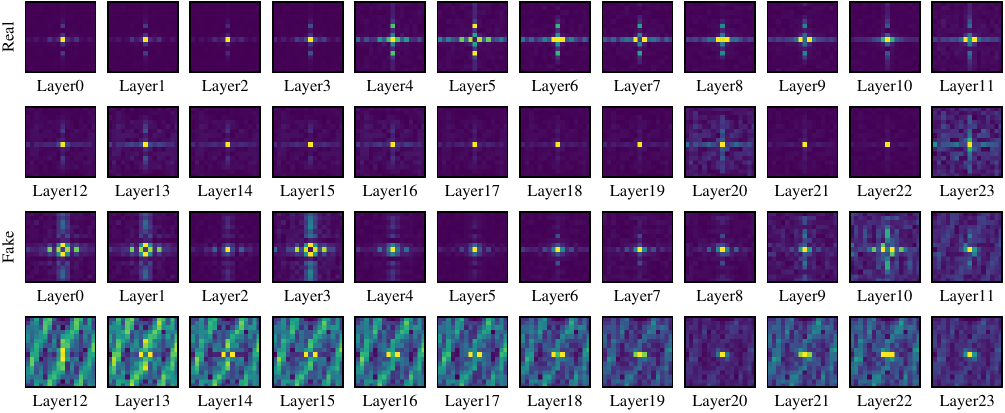}
        \caption{DFT of BigGAN.}
        \label{fig:layer_biggan_fft}
    \end{subfigure}
    \begin{subfigure}[t]{\columnwidth}
        \centering
        \includegraphics[height=2in]{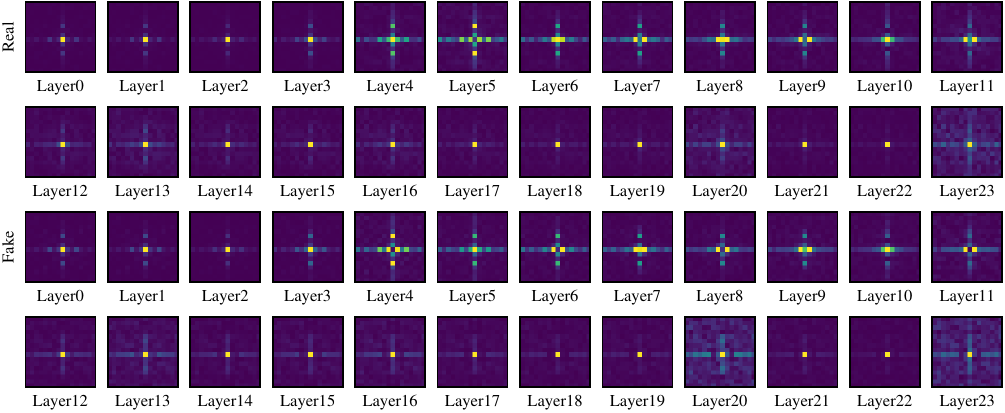}
        \caption{DFT of Wukong.}
        \label{fig:layer_wukong_fft}
    \end{subfigure}
    \begin{subfigure}[t]{\columnwidth}
        \centering
        \includegraphics[height=2in]{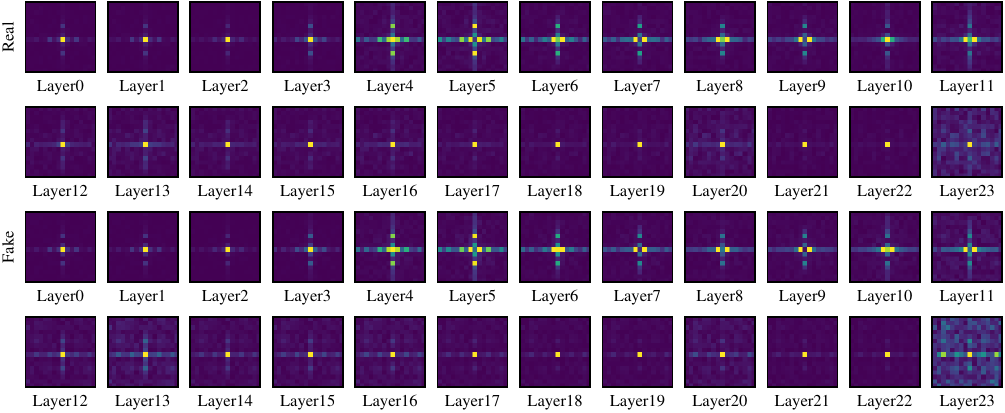}
        \caption{DFT of MidJourney.}
        \label{fig:layer_mj_fft}
    \end{subfigure}

    \caption{\textbf{DFT result plots for BigGAN, Wukong, and Midjourney.}}
    \label{fig:layers_fft1_vis}

\end{figure}

\begin{figure}[ht]
    \centering
\begin{subfigure}[t]{\textwidth}
    \centering
    \includegraphics[height=2in]{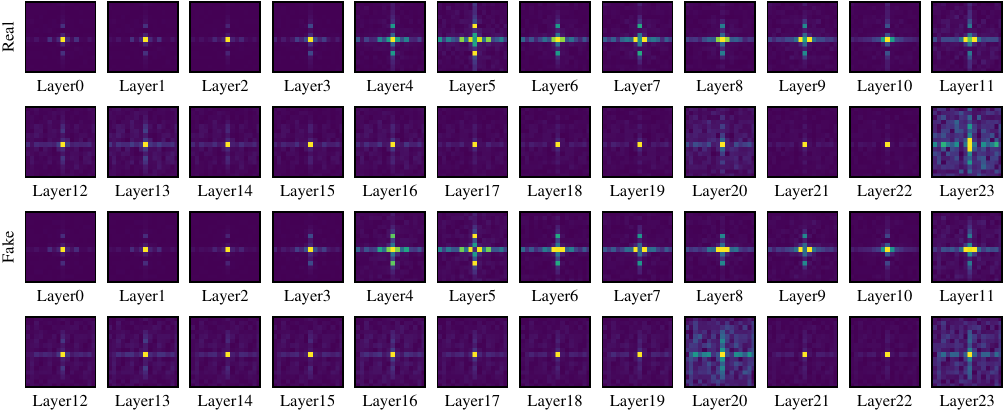}
    \caption{DFT of ADM.}
    \label{fig:layer_adm_fft}
\end{subfigure}
\begin{subfigure}[t]{\textwidth}
    \centering
    \includegraphics[height=2in]{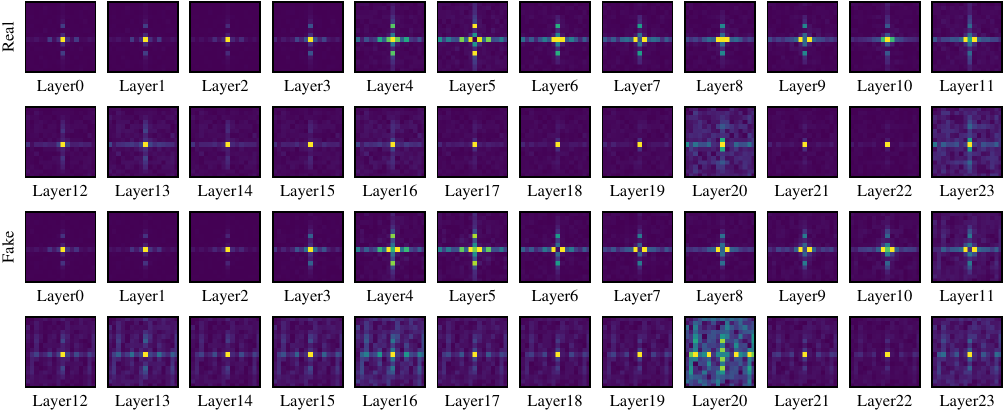}
    \caption{DFT of GLIDE.}
    \label{fig:layer_glide_fft}
\end{subfigure}
\begin{subfigure}[t]{\textwidth}
    \centering
    \includegraphics[height=2in]{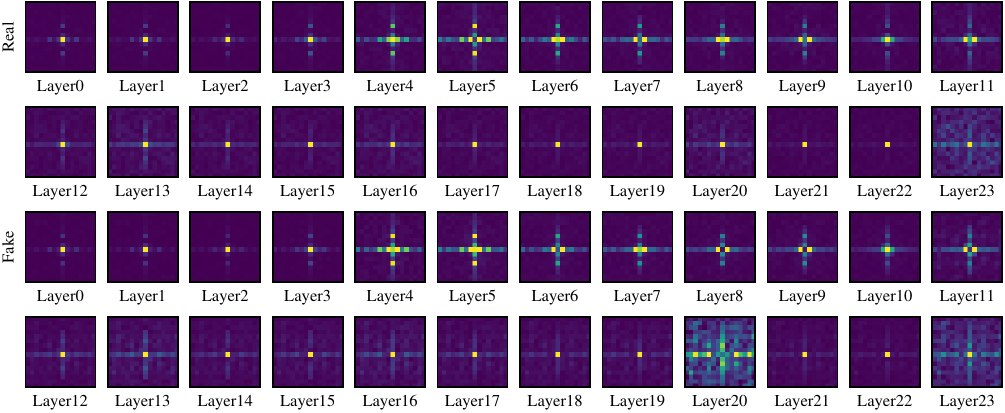}
    \caption{DFT of VQDM.}
    \label{fig:layer_vqdm_fft}
\end{subfigure}
\caption{\textbf{DFT result plots for ADM, GLIDE, and VQDM.}}
\label{fig:layers_fft2_vis}

\end{figure}

\paragraph{Discrete Fourier Transform (DFT).} To investigate why the ``bias'' phenomenon exhibits layer-wise variations, we performed Discrete Fourier Transform (DFT) on the outputs of different layers and conducted frequency-domain analysis of different layers' features. Specifically, We applied Discrete Fourier Transform (DFT) to feature maps of real and fake images encoded by the ViT, shifting the zero-frequency component to the spectrum center. Results are shown in \cref{fig:sd_layers_fft}.
DFT analysis reveals that layers 0 to 11 of the ViT primarily extract low-frequency information, such as image backgrounds. 
By Layer12, low-frequency information significantly decreases, explaining its lower difference in the ``bias'' we identified before.
We attribute Layer11's superior performance to its balance between feature extraction and ``bias'' information, retaining low-frequency data while maintaining robust feature extraction capabilities. Further more,  we conducted the same experiment as in \cref{fig:sd_layers_fft} for BigGAN, Wukong, MidJourney, ADM, GLIDE, and VQDM, with the results shown in \cref{fig:layers_fft1_vis} and \cref{fig:layers_fft2_vis}.
Among them, the spectral results of images generated by BigGAN exhibit a distinctly different pattern compared to other generative models, as previously noted in prior work \cite{cnn2020, frequency2021,unifd2023,fatformer2024}. This also indicates that the features extracted by CLIP retain this information.

\section{Supplementary on Experiments}
\label{sec:appendix_exp}
\subsection{Implementation details}
\label{sec:appendix_exp_detail}
\paragraph{Hyperparameters of EP1 and EP2.}
We employed SGD as the optimizer with an initial learning rate of $0.25$. 
A polynomial learning rate decay strategy with a power of $0.8$ was applied. 
The dimension of the hidden layer is set to $64$, the batch size was $512$, $ L_p $ is set to $512$, and $ \beta $ is set to $0.15$. We exclusively employed the training and validation sets provided by GenImage \cite{genimage2024}, utilizing both validation set accuracy and AP for hyperparameter tuning. The $\mu_e$ is set to $0.5$ and the $\Sigma_e$ is set to $0.4$ in CGP. To verify the robustness of the experiments, fixed random seeds are deliberately avoided. We employ CLIP \cite{clip2021} for feature extraction in advance and conduct training exclusively on the processed features, which significantly reduces GPU memory requirements. The entire preprocessing and training pipeline is completed on a single GTX 4090 GPU.

\subsection{Additional Evaluation Benchmark}
\label{sec:appendix_exp_aigc}
We conducted evaluations on AIGCBenchmark, with results shown in \cref{tab:aigc}. Following \citet{cospy2025}, we trained on ProGAN \cite{progan2017} and recorded the evaluation results for AP and accuracy. Notably, our method exhibits a slight gap compared to CO-SPY, which we attribute to the lacking of frequency domain. Although our method can extract generalized image forgery features, it still struggles when facing significant domain differences (e.g., face datasets). On one hand, large domain differences affect the accuracy calculation with a threshold of 0.5. For instance, as shown in the \cref{tab:aigc}, our method achieves high AP for WFIR images but only 58\% accuracy. On the other hand, relying solely on image features to generalize between face forgery and synthetic images is a formidable challenge. In contrast, extracting generalized frequency-domain features is a relatively mature solution. However, to emphasize the primary contribution of our method, we did not incorporate a frequency-domain module to pursue further performance improvements.
Nevertheless, our method still demonstrates strong generalization performance.
\begin{table*}[ht]
    \caption{\textbf{AP and Accuracy on AIGCBenchmark \cite{patchcraft2023}.} Train on CNNDet \cite{cnn2020}, and evaluate on AIGCBenchmark. Our results are reported as \textbf{Mean (STD)}, computed from 3 independent runs with different random seeds.  The baseline results are sourced from \citet{cospy2025}. The highest values in each column are highlighted in \textbf{bold} and the second-highest results are indicated with \underline{underlining}.} 

    \resizebox{\linewidth}{!}{\begin{tabular}{ccccccccccccccccccccccc}
        \toprule
         \multirow{2}{*}{Detector} & \multicolumn{2}{c}{CNNDet \cite{cnn2020}} & \multicolumn{2}{c}{FreqFD \cite{frequency2021}} & \multicolumn{2}{c}{Fusing \cite{fusing2022}} & \multicolumn{2}{c}{LNP \cite{lnp2022}} & \multicolumn{2}{c}{LGrad \cite{log2023}} & \multicolumn{2}{c}{UnivFD \cite{unifd2023}} & \multicolumn{2}{c}{DIRE-G \cite{dire2023}} & \multicolumn{2}{c}{FreqNet \cite{freqnet2024}} & \multicolumn{2}{c}{NPR \cite{npr2024}} & \multicolumn{2}{c}{CO-SPY \cite{cospy2025}} & \multicolumn{2}{c}{Ours} \\
         & AP & Acc & AP & Acc & AP & Acc & AP & Acc & AP & Acc & AP & Acc & AP & Acc & AP & Acc & AP & Acc & AP & Acc & AP & Acc \\
         \midrule
         ProGAN & \textbf{100.0} & \textbf{100.0} & \textbf{100.0} & \underline{99.86} & \textbf{100.0} & \textbf{100.0} & 99.75 & 97.31 & 99.32 & 87.78 & \textbf{100.0} & 99.81 & 91.54 & 91.80 & \textbf{100.0} & 99.58 & 99.95 & 99.84 & \textbf{100.0} & \underline{99.86} &   \textbf{100.00 (0.00)} &  99.72 (0.08)      \\
         StyleGAN & 99.19 & 72.61 & 95.81 & 86.56 & 99.26 & 82.92 & 98.55 & 92.31 & 95.53 & 77.93 & 97.48 & 80.40 & 85.18 & 71.90 & \underline{99.78} & 89.91 & 99.74 & \textbf{97.52} & \textbf{99.94} & \underline{96.29} & 98.79 (0.39)   & 87.96 (3.02)      \\
         BigGAN & 90.39 & 59.45 & 70.54 & 69.77 & 95.65 & 78.47 & 94.51 & 84.95 & 80.01 & 74.85 & \underline{99.27} & \textbf{95.08} & 74.52 & 69.10 & 96.05 & 90.45 & 84.39 & 83.20 & \textbf{99.52} & \underline{92.00} & 98.96 (0.12)   &  91.88 (3.41)       \\
         CycleGAN & 97.92 & 84.63 & 88.06 & 70.82 & 98.47 & 91.11 & 97.09 & 86.00 & 96.66 & 90.12 & \underline{99.80} & \textbf{98.33} & 71.50 & 66.80 & 99.63 & 95.84 & 97.83 & 94.10 & 99.33 & \underline{98.03} & \textbf{99.92 (0.01)}   & 94.96 (2.73)       \\
         StarGAN & 97.51 & 84.74 & \textbf{100.0} & 96.87 & 99.05 & 91.40 & 99.94 & 85.12 & 99.00 & 94.15 & 99.37 & 95.75 & 94.42 & 88.50 & 99.80 & 85.67 & \textbf{100.0} & \textbf{99.70} & \textbf{100.0} & 96.05& 99.99 (0.01)   &  96.73 (2.75)       \\
         GauGAN & 98.77 & 82.86 & 74.42 & 65.69 & 98.60 & 86.27 & 76.51 & 71.74 & 83.45 & 72.86 & \textbf{99.98} & \textbf{99.47} & 80.90 & 72.90 & 98.63 & \underline{93.41} & 81.73 & 79.97 & \underline{99.95} & 90.90& 99.93 (0.01)   &  87.90 (8.04)       \\
         StyleGAN-2 & 99.03 & 69.22 & 95.59 & 80.17 & 98.84 & 78.97 & 98.98 & 94.14 & 90.85 & 72.25 & 97.71 & 70.76 & 78.73 & 72.80 & 99.58 & 87.89 & \textbf{99.97} & \textbf{99.34} & \underline{99.94} & \underline{97.89}& 99.51 (0.14)   & 92.16 (3.77)        \\
         WFIR & 91.27 & 56.60 & 43.54 & 45.30 & \underline{95.07} & \textbf{81.95} & 74.03 & 61.80 & 70.26 & 57.30 & 94.22 & \underline{72.70} & 62.70 & 60.40 & 51.06 & 49.20 & 61.55 & 59.75 & 92.12 & 71.65& \textbf{96.56 (0.45)}   &  58.17 (2.56)       \\
         ADM & 64.70 & 51.04 & 60.30 & 61.82 & 60.26 & 51.68 & 80.78 & 71.94 & 51.92 & 55.18 & 89.80 & 67.46 & 70.00 & 64.80 & \underline{92.13} & \textbf{84.06} & 73.22 & 68.95 & \textbf{95.31} & 73.28& 91.75 (0.70)   &\underline{ 82.29 (2.79) }       \\
         Glide & 71.61 & 52.78 & 67.69 & 58.34 & 60.45 & 52.85 & 72.21 & 62.29 & 67.69 & 68.64 & 88.04 & 63.09 & 57.52 & 57.50 & 89.78 & 82.78 & 81.01 & 75.51 & \textbf{98.87} & \textbf{88.82}& \underline{95.31 (0.52)}   & \underline{88.05 (1.43)}       \\
         Midjourney & 53.45 & 50.60 & 48.71 & 46.93 & 48.78 & 50.79 & 79.54 & 70.12 & 62.77 & 58.83 & 49.72 & 49.87 & 54.62 & 53.10 & 80.88 & 71.02 & 80.33 & 74.57 &\textbf{ 89.78} & \textbf{88.70} & \underline{87.21 (0.46)}   &  \underline{76.61 (3.24)}       \\
         SD V1.4 & 55.77 & 50.14 & 43.55 & 45.01 & 52.27 & 50.13 & 63.97 & 59.29 & 66.47 & 67.24 & 68.63 & 51.70 & 52.66 & 52.40 & 77.10 & 65.56 & 80.44 & 75.58 & \underline{93.30} &\underline{ 86.01}& \textbf{94.30 (0.91)}   &\textbf{86.76 (2.79)}        \\
         SD V1.5 & 55.68 & 50.07 & 43.09 & 44.27 & 51.99 & 50.07 & 64.16 & 59.26 & 65.91 & 66.40 & 68.07 & 51.59 & 53.17 & 53.00 & 77.95 & 65.84 & 81.23 & 76.36 &\underline{ 93.20} & \underline{86.35}& \textbf{94.37 (0.87) }  & \textbf{86.62 (2.81) }       \\
         VQDM & 72.62 & 52.15 & 69.45 & 65.34 & 71.55 & 53.93 & 66.82 & 62.70 & 54.30 & 56.92 & \underline{97.53} & \underline{86.01} & 65.87 & 58.70 & 90.42 & 82.29 & 74.66 & 72.98 & \textbf{97.73} & 82.35& 94.77 (0.51)   & \textbf{86.48 (1.66)   }     \\
         wukong & 52.71 & 50.08 & 46.96 & 48.48 & 51.53 & 50.13 & 61.99 & 56.75 & 69.84 & 68.41 & 78.44 & 55.14 & 51.86 & 48.70 & 69.43 & 58.59 & 75.42 & 72.23 & \underline{92.34} & \underline{78.77}& \textbf{94.55 (0.79) }  &  \textbf{86.16 (3.02)}       \\
         DALLE 2 & 47.16 & 49.85 & 39.58 & 36.00 & 37.89 & 49.50 & 87.88 & 76.25 & 64.21 & 57.45 & 66.06 & 50.80 & 52.85 & 51.40 & 55.40 & 55.75 & 70.86 & 61.40 & \textbf{96.23} & \underline{77.05}& \underline{91.08 (0.59) }  & \textbf{79.17 (5.53)}        \\
         Average & 77.99 & 63.55 & 67.96 & 63.83 & 76.23 & 68.76 & 82.29 & 74.50 & 76.14 & 70.39 & 87.13 & 74.25 & 68.63 & 64.61 & 86.10 & 78.61 & 83.90 & 80.69 & \textbf{96.72} & \textbf{87.75} & \underline{96.06 (0.39) }  &  \underline{86.35 (0.67) }      \\
        \bottomrule
        \end{tabular}}
        \label{tab:aigc}
\end{table*}
\subsection{Additional Evaluation Metrics}
\label{sec:appendix_exp_metrics}

\paragraph{Real accuracy, fake accuracy, AP, F1 score and FPR95.} Here we provide  additional evaluation metrics for EP1. We computed the accuracy on real images, the accuracy on fake images, Average Precison (AP), AUROC and FPR95 metrics for our method and present them in \cref{tab:ep1_other_metrics}.
\begin{table*}[ht]
    \caption{ \textbf{More metrics on EP1 setting.} We report accuracy on real images (Real Acc), accuracy on fake images (Fake Acc), F1 score and FPR95 on EP1 setting. Train on GenImage/SDV1.4, and evaluation on GenImage test set. The experimental results are statistically derived from 3 runs with random seeds. All results are reported as \textbf{Mean (STD)}, computed from 3 independent runs with different random seeds. }
    \centering
    \resizebox{\columnwidth}{!}{\begin{tabular}{cccccccccc}
        \toprule
           Method                   & BigGAN & MidJourney & SD V1.4 & SD V1.5 & VQDM  & Wukong & GLIDE & ADM   & Avg.  \\
        \midrule
        Real Acc (\%)        & 97.54 (0.81) & 97.64 (0.87)      & 97.41 (0.93)  & 97.15 (0.79)   & 97.59 (0.76) & 97.44 (0.79)  & 97.65 (0.87) & 97.73 (0.93) & 97.52 (0.84) \\
        Fake Acc (\%)         & 99.99 (0.02) & 91.00 (3.84)      & \textbf{100.00 (0.00)}  & 99.89 (0.00)   & 99.99 (0.02) & \textbf{100.00 (0.00)}  & 99.02 (0.75) & 95.46 (2.42) & 98.11 (0.92) \\
        AP             & 99.88 (0.01)  & 98.98 (0.19)      & \textbf{100.00 (0.00)}  & 99.99 (0.00)  & 99.99 (0.00) & \textbf{100.00 (0.00)}  & 99.82 (0.04) & 99.54 (0.14) & 99.77 (0.05) \\
        \midrule
        F1 Score       & 0.99 ($2.31\times10^{-3}$) & 0.94  (0.01)      & 0.99 ($4.52\times 10^{-3}$)  & 0.99 ($3.84\times 10^{-3}$)   & 0.99 ($3.61\times10^{-3} $) & 0.99 ($3.86\times10^{-3}$)  & 0.98 ($1.02\times10^{-3}$) & 0.97 ($ 8.67\times10^{-3}$) & 0.98 ($5.66\times10^{-3}$) \\
         FPR95  (\%)         & 0.38 (0.05)  & 4.89 (1.43)      & 0.00 (0.00)  & 0.00 (0.00)  & 0.06 (0.02) & 0.00 (0.00) & 0.37 (0.08) & 1.86 (0.64) & 0.94 (0.28) \\
        \bottomrule
    \end{tabular}}
    \label{tab:ep1_other_metrics}
\end{table*}

\subsection{Robustness}
\label{sec:appendix_robustness}
\begin{figure}[ht]
    \centering
    \includegraphics[width=\linewidth]{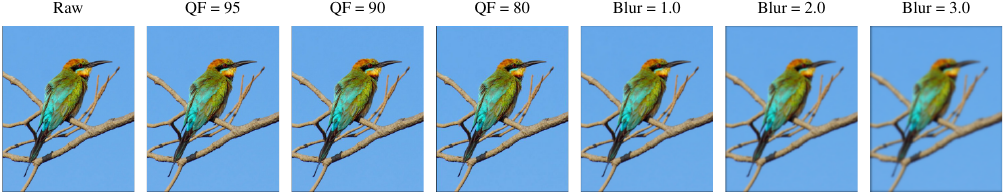}
    \caption{\textbf{Images with varying degrees of degradation processing.}}
    \label{fig:robust_eample}
\end{figure}

We subjected the GenImage \cite{genimage2024} test images to common degradation operations, including varying intensities of JPEG compression and Gaussian blur, as depicted in \cref{fig:robust_eample}. The accuracy and average precision (AP) were computed and reported as \textbf{Mean (STD)} values based on three independent runs with different random seeds (\cref{tab:exp_robust}). 

Notably, our method demonstrates significant degradation robustness in terms of AP, but exhibits suboptimal performance when evaluated by accuracy. We attribute this discrepancy to our use of Layer11 outputs as feature inputs, where the ``bias'' phenomenon is most pronounced. As reported in \citet{emergence2018}, deeper layers in well-trained neural networks typically exhibit greater robustness to perturbations. To validate this hypothesis, we conducted additional experiments using the final layer (Layer23) as input. The results confirm that Layer23 indeed shows significantly less performance degradation under image perturbations, particularly resistant to Blur degradation.

\begin{table*}[ht]
  \caption{\textbf{Accuracy and AP under common degradations.} We performed image degradation tests at varying levels on the GenImage \cite{genimage2024} test set and recorded the average accuracy at 0.5 and AP, as reported in \cref{tab:exp_robust}. Notably, we applied corresponding degradation to \textbf{all} test images. All results are reported as \textbf{Mean (STD)}, computed from 3 independent runs with different random seeds. \textbf{\dag}: the results are evaluated by their pretrained weights.} 
  \centering
  \resizebox{\columnwidth}{!}{\begin{tabular}{@{}ccccccccc@{}}
      \toprule 
      &     &     \multicolumn{4}{c}{JPEG}        &        \multicolumn{3}{c}{Blur}                \\
      \cmidrule(lr){3-6}  \cmidrule(lr){7-9}
                               &      & QF = 95      & QF = 90      & QF = 80      & QF =70 & Sigma = 1    & Sigma = 2    & Sigma = 3    \\
      \midrule
      \multirow{2}{*}{AIDE $^\dag$ \cite{aide2024}} & Acc & 59.41 (--) &  56.59 (--) & 57.80 (--) & 60.16 (--) & 68.46 (--) & 63.23 (--) & 51.07 (--) \\
                               & AP   & 85.69 (--) & 82.37 (--) & 84.33 (--) & 88.78 (--) & 75.23 (--) & 61.79 (--)        & 63.16 (--)  \\
      \midrule
      \multirow{2}{*}{Ours (Layer11)} & Acc & 92.61 (1.43) & 75.52 (5.15) & 60.51 (5.50) & 58.18 (3.72) & 58.23 (2.85) & 56.84 (0.69) & 57.81 (3.29) \\
                               & AP   & 93.47 (0.10) & 94.13 (0.23) & 90.40 (0.19) & 88.60 (0.27) & 76.43 (0.27) & 59.94 (0.29)        & 62.43 (0.41)  \\
      \multirow{2}{*}{Ours (Layer23)} & Acc & 80.84 (0.39) & 77.23 (0.39) & 71.71 (1.61) & 69.03 (1.84) & 66.17 (0.56) &      57.64 (0.85) &   52.56 (0.48)  \\
                               & AP   &  86.46 (0.06) & 83.37 (0.12) & 78.63 (0.12) & 77.06 (0.10) & 74.35 (0.08) &     66.35 (0.04) &  61.98 (0.08) \\
      \bottomrule
    \end{tabular}}
  \label{tab:exp_robust}
\end{table*}

\subsection{Additional Ablation Studies}
\label{sec:appendix_exp_aas}

\paragraph{Different Text Guidance.}
We conducted ablation experiments on $ t $ in $ r(z \mid t, y) $, testing image-paired prompts, category prompts, random characters and our proposed DTO, with results in \cref{tab:text_guidance_ablation}. 
Surprisingly, using image-paired prompts for each image performed poorly, despite separating real and fake images for conditional information. We attribute this to prompt variability, causing large discrepancies in mapped text features and complicating optimization. In contrast, category prompts achieved better performance, validating our explanation. Fixed random character strings, as expected, outperformed both semantic-based approaches. Dataset-wide pooling showed minimal improvement (0.13\%) over random text, likely due to semantic information loss. Our method, balancing mini-batch semantics and global ``bias'', achieved a 1.72\% performance gain over pooling random text features.
\begin{table*}[ht]
    \centering
    \caption{\textbf{Ablation experiments on text guidance.} “Random”: randomly generated characters;  ``An image of [class]'': [class] is ``real'' or ``fake''. All results are reported as \textbf{Mean (STD)}, computed from 3 independent runs with different random seeds. We both provide accuracy and Diff in the table.}
    \begin{tabular}{ccc}
    \toprule
    Guidance Type          & Acc(\%)   & Diff (\%)      \\
    \midrule
    ``An image of [class].'' &  96.53 (0.30)  & $\downarrow$ \textcolor{red}{-1.29}  \\
    Random     &   95.52 (0.72) & $\downarrow$ \textcolor{red}{-2.30}  \\
    Image-paired  Prompts    &  94.38 (1.59)  & $\downarrow$ \textcolor{red}{-3.44}  \\
    \rowcolor{gray!20}DTO                     &  97.82 (0.15)  & -  \\
    \bottomrule
    \end{tabular}
    \label{tab:text_guidance_ablation}
\end{table*}
\paragraph{Dimensionality reduction compression methods.} We conducted ablation experiments on different dimensionality reduction methods for text processing branch, employing PCA, linear projection, and Top-K absolute values of vector components as reduction approaches, with results presented in \cref{tab:ablation_ld}.
\begin{table*}[ht]
  \caption{\textbf{Ablation experiments on different text dimension reduction methods.} All results are reported as \textbf{Mean (STD)}, computed from 3 independent runs with different random seeds.}
  \centering
    \resizebox{\columnwidth}{!}{\begin{tabular}{cccccccccc}
        \toprule
        Method & BigGAN                   & MidJourney   & SD V1.4      & SD V1.5      & VQDM         & Wukong       & GLIDE        & ADM                   & Avg.                  \\
        \midrule
        PCA    &  93.65 (2.70) & 82.27 (1.80) & {99.86 (0.01)} & {99.66 (0.02)} & {96.19 (0.81)} & {99.02 (0.14)} & 97.36 (1.11) & 90.60 (1.76)          & 94.83 (1.01)      \\
        Top-K    & 93.61 (2.24)             & 81.91 (2.02) & \textbf{99.89 (0.04)} & \textbf{99.71 (0.01)} & \textbf{96.21 (1.21)} & \textbf{99.13 (0.19)} & 96.62 (0.49) & 90.18 (1.49) & 94.66 (0.92)\\
        Linear & \textbf{98.55 (0.22)}             & \textbf{94.32 (1.54)} & 98.71 (0.46) & 98.52 (0.39) & 94.79 (0.37) & 98.72 (0.39) &\textbf{ 98.34 (0.11)} & \textbf{96.59 (0.80)} & \textbf{97.82 (0.15)} \\
        \bottomrule
      \end{tabular}}
    \label{tab:ablation_ld}
\end{table*}

\subsection{Additional Visualizations}
\label{sec:appendix_exp_vis}

\begin{figure}[hb]
    \centering
    \begin{subfigure}[t]{0.31\columnwidth}
        \centering
        \includegraphics[width=\columnwidth]{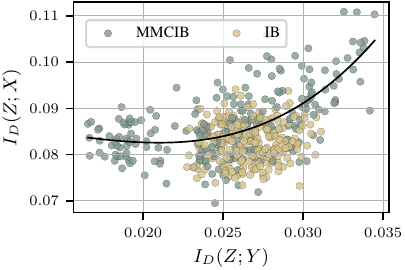}
        \caption{ VQDM.}
    \end{subfigure}
    \begin{subfigure}[t]{0.31\columnwidth}
        \centering
        \includegraphics[width=\columnwidth]{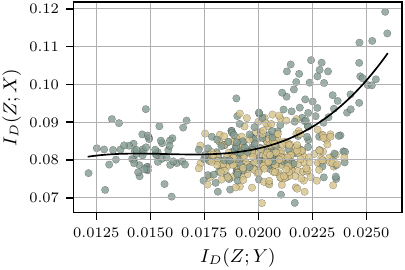}
        \caption{ MidJourney.}
    \end{subfigure}
    \centering
    \begin{subfigure}[t]{0.31\columnwidth}
        \centering
        \includegraphics[width=\columnwidth]{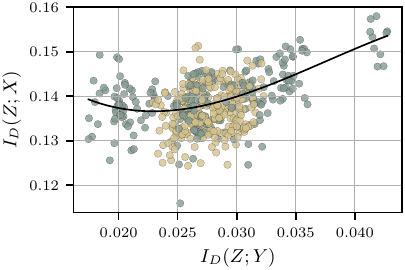}
        \caption{ GLIDE.}
    \end{subfigure}

    \begin{subfigure}[t]{0.31\columnwidth}
        \centering
        \includegraphics[width=\columnwidth]{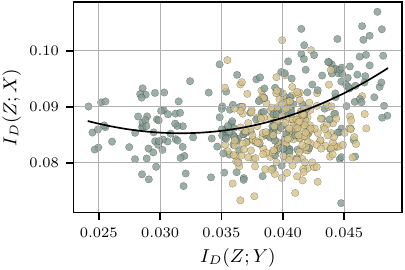}
        \caption{ SDV1.5.}
    \end{subfigure}
    \begin{subfigure}[t]{0.31\columnwidth}
        \centering
        \includegraphics[width=\columnwidth]{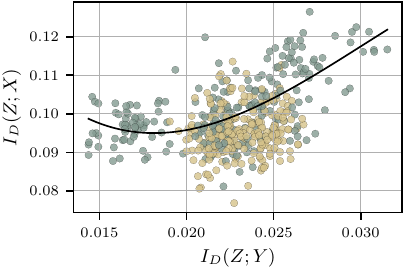}
        \caption{ ADM.}
    \end{subfigure}
    \begin{subfigure}[t]{0.31\columnwidth}
        \centering
        \includegraphics[width=\columnwidth]{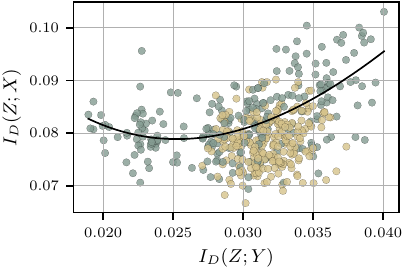}
        \caption{ Wukong.}
    \end{subfigure}
    \caption{Comparison of $I_D(X;Z)$ against $I_D(Z;Y)$ on \textbf{M}ulti\textbf{M}odal \textbf{C}onditional \textbf{I}nformation \textbf{B}ottleneck (MMCIB) and \textbf{I}nformation \textbf{B}ottleneck (IB). The \textbf{black line} represents the trend of MMCIB obtained through triple curve fitting.}
    \label{fig:more_mie}
\end{figure}

\paragraph{Mutual Information Estimation.} We provide more visualization results same as \cref{fig:izx_izy}. The results are shown in \cref{fig:more_mie}, where each subclass consistently aligns with the conclusions we derived in \cref{sec:5_5_mie}.

\paragraph{Discrete Fourier Transform.}
We conduct more visualization of Discrete Fourier Transform experiments same as \cref{sec:appendix_layers} on BigGAN, Wukong, MidJourney, ADM, GLIDE and VQDM. All of them are sourced from GenImage \cite{genimage2024}.
\label{sec:appendix_exp_dft}

\begin{figure}[ht]
    \centering
    \begin{subfigure}[t]{0.23\textwidth}
        \centering
        \includegraphics[width=\textwidth]{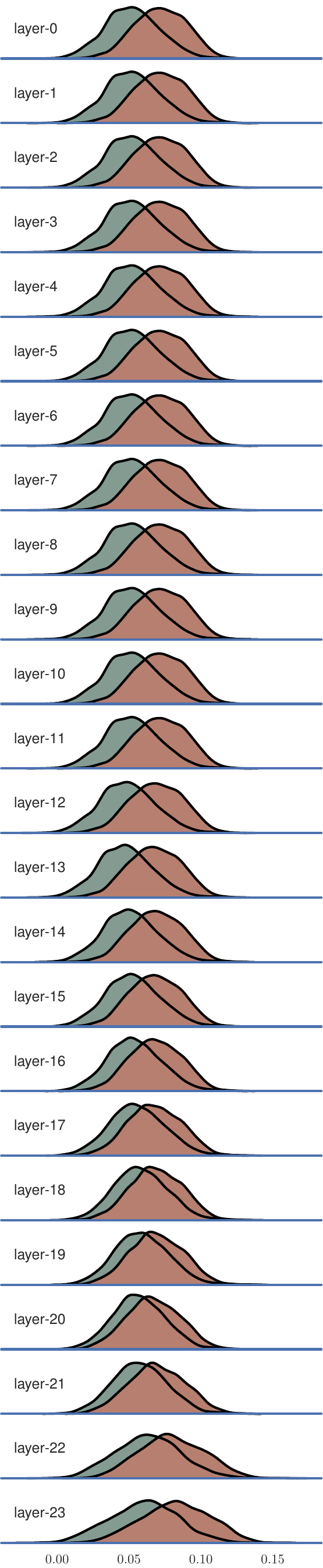}
        \caption{Wukong.}
        \label{fig:layer_wukong}
    \end{subfigure}
    \begin{subfigure}[t]{0.23\textwidth}
        \centering
        \includegraphics[width=\columnwidth]{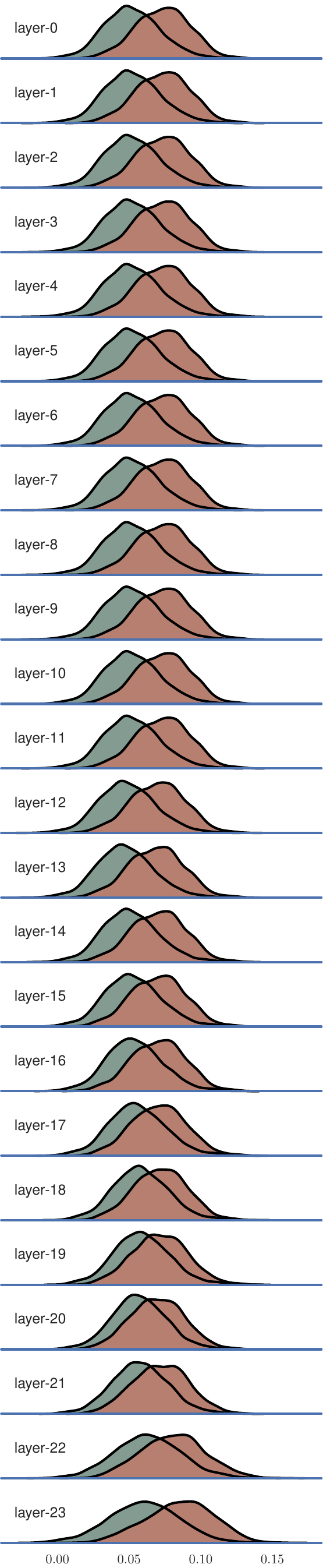}
        \caption{VQDM.}
        \label{fig:layer_vqdm}
    \end{subfigure}
    \begin{subfigure}[t]{0.23\textwidth}
        \centering
        \includegraphics[width=\textwidth]{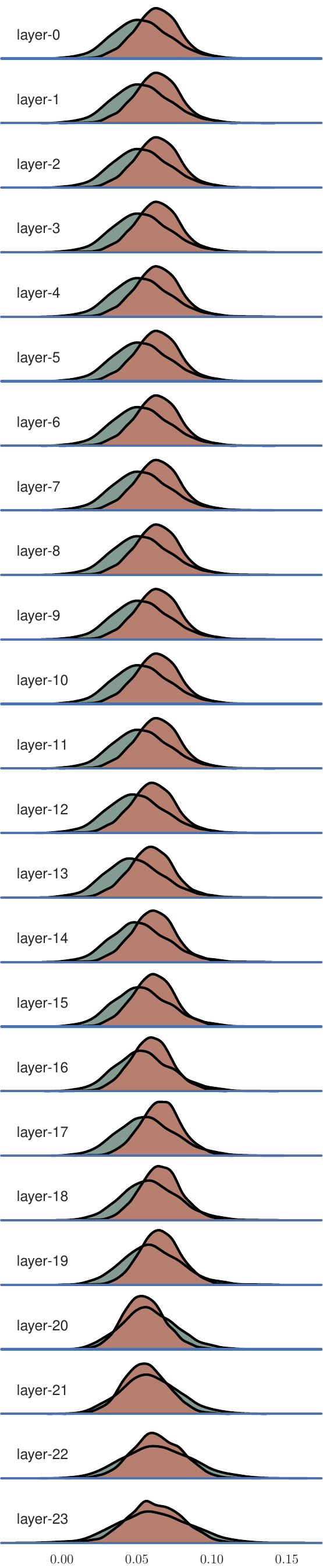}
        \caption{GLIDE.}
        \label{fig:layer_glide}
    \end{subfigure}
    \begin{subfigure}[t]{0.23\textwidth}
        \centering
        \includegraphics[width=\textwidth]{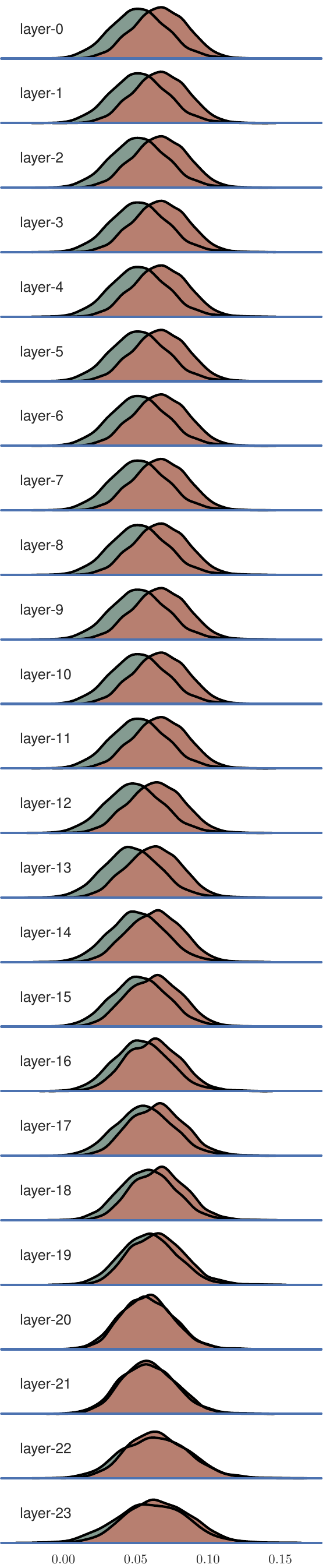}
        \caption{ADM.}
        \label{fig:layer_adm}
    \end{subfigure}
    \caption{\textbf{Visualization of cosine similarity between the features of real and fake images generated by Wukong, VQDM, GLIDE, and ADM models, and random text features, across different layers of the ViT in CLIP.} The features were transformed using the projection layer of the ViT. The real images were sourced from ImageNet, and there were 2000 fake images and 2000 real images. Notably, among these four sets of experiments, the results on the ADM dataset are the most pronounced. The cosine similarity distributions between real and fake images for the pooled text are nearly identical at the final layer, whereas the difference is relatively more significant at layer11. Our method utilizes the output of layer11, which is why the accuracy for the ADM model is significantly higher compared to other methods.}
    \label{fig:layers_hist_vis}
\end{figure}

\paragraph{The frequency distribution of cosine similarity between different layers and random text features.}
We tested the differences in cosine similarity between the [CLS] output features of different layers of CLIP's ViT, after projection, and random text features on various datasets, with the results shown in \cref{fig:layers_hist_vis}.
In the GLIDE and ADM datasets, the differences in ``bias'' among different layers of CLIP's ViT are quite pronounced. 
It is evident that the cosine similarity distributions between real and fake images and random text features gradually converge, indicating that the discriminative ability of ``bias'' diminishes in the later layers. 
While the Wukong and VQDM datasets do not exhibit as stark a contrast as the former two, the shape of their distributions suggests that the ``bias'' discriminative ability also decreases in the later layers of Wukong and VQDM.

\end{document}